\documentclass[runningheads]{llncs}

\usepackage{eccv}

\usepackage{eccvabbrv}

\usepackage{graphicx}
\usepackage{booktabs}

\usepackage[accsupp]{axessibility}  

\usepackage[pagebackref,breaklinks,colorlinks,citecolor=eccvblue]{hyperref}

\usepackage{orcidlink}

\begin{document}

\title{DreamDissector: Learning Disentangled Text-to-3D Generation from 2D Diffusion Priors} 

\titlerunning{DreamDissector}

\author{Zizheng Yan\inst{123} \and
Jiapeng Zhou\inst{123} \and
Fanpeng Meng\inst{123} \and
Yushuang Wu\inst{123}  \and
Lingteng Qiu\inst{123} \and
Zisheng Ye\inst{123} \and
Shuguang Cui\inst{213} \and
Guanying Chen\inst{13} \and
Xiaoguang Han\inst{213}\thanks{Corresponding Author.} 
}

\authorrunning{Z.~Yan et al.}

\institute{
\mbox{Shenzhen Future Network of Intelligence Institute \hspace{5mm}  \and SSE, CUHKSZ} \and
\mbox{Guangdong Provincial Key Laboratory of Future Networks of Intelligence, CUHKSZ} \\
\email{zizhengyan@link.cuhk.edu.cn}
}

\maketitle

{
    \centering
    \captionsetup{type=figure}
    \includegraphics[width=\textwidth]{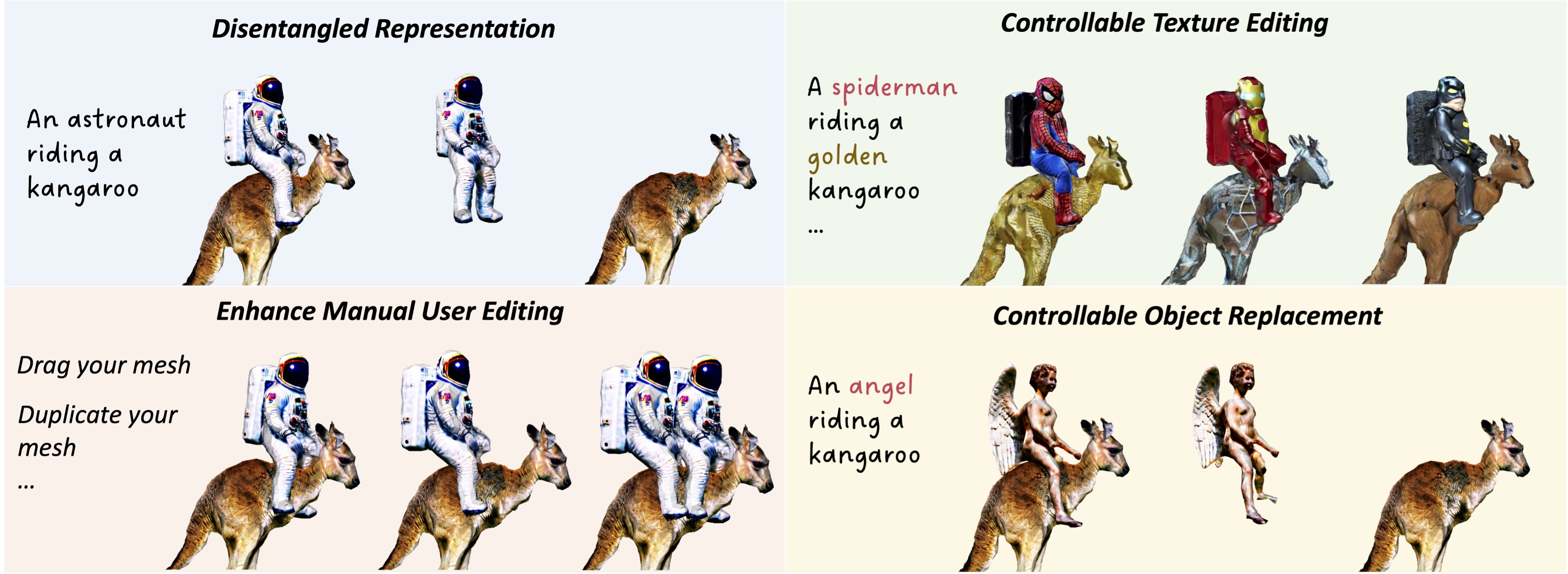}
    \captionof{figure}{
    \textbf{Results and applications of our DreamDissector.} \textbf{Top left}: DreamDissector can generate multiple independent textured meshes with plausible interactions. \textbf{Top right}: Facilitating text-guided texturing at the object level. \textbf{Bottom left}: Enhancing convenient manual user geometry editing through simple operations. \textbf{Bottom right}: Facilitating text-guided controllable object replacement.
    }
    \label{fig:teaser}
}
\begin{abstract}
Text-to-3D generation has recently seen significant progress. To enhance its practicality in real-world applications, it is crucial to generate multiple independent objects with interactions, similar to layer-compositing in 2D image editing. However, existing text-to-3D methods struggle with this task, as they are designed to generate either non-independent objects or independent objects lacking spatially plausible interactions. Addressing this, we propose DreamDissector, a text-to-3D method capable of generating multiple independent objects with interactions. DreamDissector accepts a multi-object text-to-3D NeRF as input and produces independent textured meshes. To achieve this, we introduce the Neural Category Field (NeCF) for disentangling the input NeRF. Additionally, we present the Category Score Distillation Sampling (CSDS), facilitated by a Deep Concept Mining (DCM) module, to tackle the concept gap issue in diffusion models. By leveraging NeCF and CSDS, we can effectively derive sub-NeRFs from the original scene. Further refinement enhances geometry and texture. Our experimental results validate the effectiveness of DreamDissector, providing users with novel means to control 3D synthesis at the object level and potentially opening avenues for various creative applications in the future.
  \keywords{Text-to-3D \and Disentangled Generation \and Diffusion Prior}
\end{abstract}

\section{Introduction}
\hspace{10pt} 
Creating high-quality 3D content is crucial for enhancing the visual experience in movies, video games, and emerging augmented/virtual reality environments. With the advent of deep learning, numerous techniques for generating 3D content have been proposed~\cite{luo2021diffusion,nichol2022point,zhou20213d,zeng2022lion,henzler2019escaping,lunz2020inverse,gao2022get3d,wu2023scoda}. Despite this, the pace of progress in 3D content generation has been relatively slow compared to that of 2D image generation. Recently, text-to-image (T2I) generation has seen remarkable progress~\cite{imagen,rombach2022high,ramesh2022hierarchical}, propelled by advances in diffusion models~\cite{ho2020denoising,song2019generative} and large-scale training data~\cite{schuhmann2022laion,schuhmann2021laion}. Observing the success of T2I models, researchers have begun adapting the knowledge from T2I models for 3D generation. Poole~\etal~\cite{poole2022dreamfusion} introduced a method known as Score Distillation Sampling (SDS), which optimizes a Neural Radiance Field (NeRF) using the T2I diffusion model for guidance, attracting significant interest in the 3D generation domain. 
 
Among these advancements, there has been research focused on enhancing the practicality of 2D content creation in real-world applications, such as layered image generation~\cite{bar2022text2live,zhang2023text2layer}. 
Inspired by this, the concept of ``layered'' objects capable of interacting in a 3D space is an emerging area of interest for practical applications. Consider the task of building a 3D model where an astronaut is riding a kangaroo, as shown in Figure~\ref{fig:teaser}. The ability to individually manipulate each element, such as reorienting the kangaroo or modifying the kangaroo's texture without impacting the astronaut's model, offers significant convenience and flexibility, thereby greatly enhancing the creative workflow for human artists. To facilitate such precise editing, the objects should be independently represented, meaning they should have separate surface meshes at the object level.

However, existing text-to-3D methods struggle to fulfill such a complex task as they are designed to generate either non-independent objects or an assembly of independent objects that lack spatially plausible interactions. CompoNeRF~\cite{lin2023componerf} and Comp3D~\cite{po2023compositional} utilize a set of 3D bounding boxes as input conditions for score distillation sampling to generate multiple independent and composable NeRFs. Although the generated local NeRFs can be composed into a global scene, the requirement for 3D input bounding boxes limits the objects to having only simple interactions, \eg, a bed positioned next to a nightstand. Consequently, these methods struggle to generate objects with complex interactions, \eg, an astronaut riding a kangaroo.

To this end, we propose DreamDissector, a text-to-3D method that generates multiple independent objects with interactions. DreamDissector begins with a text-to-3D NeRF containing multiple interacting objects and produces several independent textured meshes, maintaining the interactions and overall appearance while improving geometries and textures. This process comprises two phases: disentanglement and refinement.
Specifically, we introduce a novel Neural Category Field (NeCF) representation that learns a probability distribution for each point in space across all categories. This allows the original density field to be decomposed according to the distribution, enabling the input NeRF to be disentangled into multiple independent sub-NeRFs for each category.
For training the disentanglement phase, we introduce a Category Score Distillation Sampling~(CSDS) loss, which consists of a set of SDS losses for the subjects depicted in the input NeRF. However, we observed that the CSDS with vanilla diffusion model is inadequate for addressing concept gaps --- discrepancies where the object generated from the complete text prompt and the category-specific text prompt occupy different areas in the T2I diffusion model's latent space. This leads to mismatched disentanglement, as shown in Figure~\ref{fig:discrepancy_dcm} left.
To overcome these concept gaps, we leverage a technique called Deep Concept Mining~(DCM), which personalizes the diffusion model by learning the profound concept portrayed by the input NeRF. This personalized diffusion model significantly enhances the disentanglement phase.
Following disentanglement, we transform the sub-NeRFs into DMTets~\cite{dmtet} for further refinement, which not only corrects certain artifacts, resulting in improved geometry and texture, but also enables the export of multiple independent surface meshes. Finally, we demonstrate the practicality of DreamDissector through a set of controllable editing applications.

To summarize, our contributions are as follows:
\begin{itemize}
    \item To the best of our knowledge, we are the first to address the problem of disentangling text-to-3D NeRFs.
    \item To address the problem, we introduce a novel framework named DreamDissector, including a novel Neural Category Field (NeCF) representation that can disentangle an input NeRF into indepedent sub-NeRFs, a Deep Concept Mining (DCM) technique to facilitate the alignment between sub-NeRFs and concepts through personalizing a diffusion model, and a Category Score Distillation Sampling (CSDS) loss that harnesses DCM to enhance the NeCF learning.
    \item Experimental results demonstrate the effectiveness of DreamDissector, and additional controllable editing applications illustrate its practicality in real scenarios.
\end{itemize}

\section{Related Works}
\noindent \textbf{3D Generative models.}
Notable advancements have been witnessed in 3D generative models in recent years. Among those advancements, various representations have been explored, including point clouds~\cite{nichol2022point,zhou20213d,zeng2022lion,achlioptas2018learning,mo2019structurenet,yang2019pointflow}, 3D voxels~\cite{henzler2019escaping,lunz2020inverse, smith2017improved,wu2016learning}, meshes~\cite{gao2022get3d,zhang2020image}, and implicit representations~\cite{chen2019learning,mescheder2019occupancy}. However, a substantial performance gap exists between these 3D generative models and their 2D counterparts, primarily attributable to the lack of 3D training data. Despite several endeavors to develop 3D-aware generative models from 2D image collections, generating high-fidelity assets with 3D consistency and scalability is still challenging.
%

\noindent \textbf{Text-to-3D Generation.}
On the other hand, T2I generation thrives on the huge amount of 2D image data. Therefore, researchers intuitively tried to make use of powerful T2I models for later 3D generations. One of the representative methods is to use the CLIP model~\cite{radford2021learning}, which can match input texts and rendered images better than any mechanisms before. Some early text-to-3D work directly employed CLIP to supervise 3D features during generation, such as CLIP-Mesh~\cite{Mohammad_Khalid_2022} and Text2Mesh~\cite{michel2021text2mesh} focusing on mesh, and so on~\cite{jain2022zeroshot,lee2022understanding,hong2022avatarclip}.
Similar to CLIP, the Diffusion model~\cite{rombach2022high,ramesh2022hierarchical,wang2022pretraining} caught researchers' attention for its promising performance in T2I. Poole\etal~\cite{poole2022dreamfusion} proposed a novel method of Score Distillation Sampling (SDS) to optimize 3D models from their multiple views generated by a 2D diffusion model. Concurrently, Wang~\etal~\cite{wang2022score} proposed a score jacobian chaining method to accomplish similar tasks. However, these methods still suffer from multi-view inconsistency problems. To improve the 3D synthesis performance, follow-up research can be roughly classified into several categories --- pursuing more effective expressions of SDS~\cite{raj2023dreambooth3d,metzer2022latentnerf,wang2023prolificdreamer,hong2023debiasing,huang2023dreamtime}, applying two-stage coarse-to-fine optimization~\cite{lin2023magic3d,chen2023fantasia3d}, introducing more 3D priors or constructing 3D datasets~\cite{seo2023let,melaskyriazi2023realfusion,liu2023zero1to3,shi2023mvdream,liu2023syncdreamer,yu2023mvimgnet}, and utilizing the power of more models such as CLIP~\cite{wang2022rodin,xu2023bridging} and LLMs~\cite{yang2023llmgrounder,hong20233dllm}.

\noindent \textbf{Multi-object 3D Generation.}
For multi-object text-to-3D generation, most methods model the objects in a scene collectively, as seen in Text2NeRF~\cite{zhang2023text2nerf}, Text2Room~\cite{höllein2023text2room}, and SceneScape~\cite{fridman2023scenescape}. Compared to earlier methods, these can create more photorealistic scenes with intricate geometry and textures. Although this reduces complexity in the design of modeling algorithms, treating multiple objects as a single undivided scene limits the ability to edit each object individually. Some recent work~\cite{lin2023componerf,po2023compositional} has attempted to simultaneously generate multiple composable 3D objects. However, these methods require additional 3D bounding boxes as input to specify the objects' positions within a scene, which increases the user's workload. 
There are also some concurrent works addressing the similar problem~\cite{zhou2024gala3d,gao2024graphdreamer,epstein2024disentangled}.
\section{Preliminary}
\noindent \textbf{Neural Radiance Field}~(NeRF)~\cite{mildenhall2021nerf} is a representation proposed to synthesize the novel views of a 3D scene. NeRF employs a multi-layer perception (MLP) to directly model the density $\sigma$ and the color $\mathbf{c}$ at each point in a 3D space. Given a camera ray $\mathbf{r}(t) = \mathbf{o} + t\mathbf{d}$ defined by origin $\mathbf{o}$ and the view direction $\mathbf{d}$, the input points are sampled along the ray at different depth $t$, and the colors $\mathbf{c}_i$ and densities $\sigma_i$ are obtained by the MLP. Then, volume rendering is used to compute the final color of the pixels:
\begin{align}
C(\mathbf{r}) &= \sum_{i=1}^{N} \alpha_i(1 - \exp(-\sigma_i\delta_i))\mathbf{c}_i
\end{align}
where $\alpha_i = \exp(-\sum_{j=1}^{i-1}\sigma_j\delta_j)$ denotes the transmittance probability at $i$,
 and $\delta_i$ denotes the distance between consecutive samples.

\noindent \textbf{Score Distillation Sampling}~(SDS) is the key to the success of DreamFusion~\cite{poole2022dreamfusion}, which aims to distill  the knowledge from a pre-trained text-to-image diffusion-based generative model~\cite{rombach2022high,imagen} to achieve text-to-3D generation. The diffusion model consists of a denoising function $\epsilon_{\phi}(x_t, y, t)$~\cite{ho2020denoising} that predicts the sampled noise $\epsilon$ given the noisy image $x_t$, the time step $t$, and text prompt embedding $y$. By feeding a set of rendered images of the NeRF parameterized by $\mathbf{\theta}$ to the frozen denoising function $\epsilon_{\phi}(x_t, y, t)$, the NeRF is optimized towards the direction where the rendered images closer to the high probability density regions conditioned on the text embedding under the 2D diffusion prior. Specifically, the gradients of SDS are estimated as follows,
\begin{equation}
\nabla_\theta L_{SDS}(\phi, \theta) = \mathbb{E}_{t, \epsilon} \left[ w(t) \left( \epsilon_\phi(x_t; y, t) - \epsilon \right) \frac{\partial x}{\partial \theta} \right],
\end{equation}
where $w(t)$ is the weighting function controlling the strength of the SDS guidance.

\section{Methodology}
\begin{figure*}[t]
    \centering
    \includegraphics[width=0.85\linewidth]{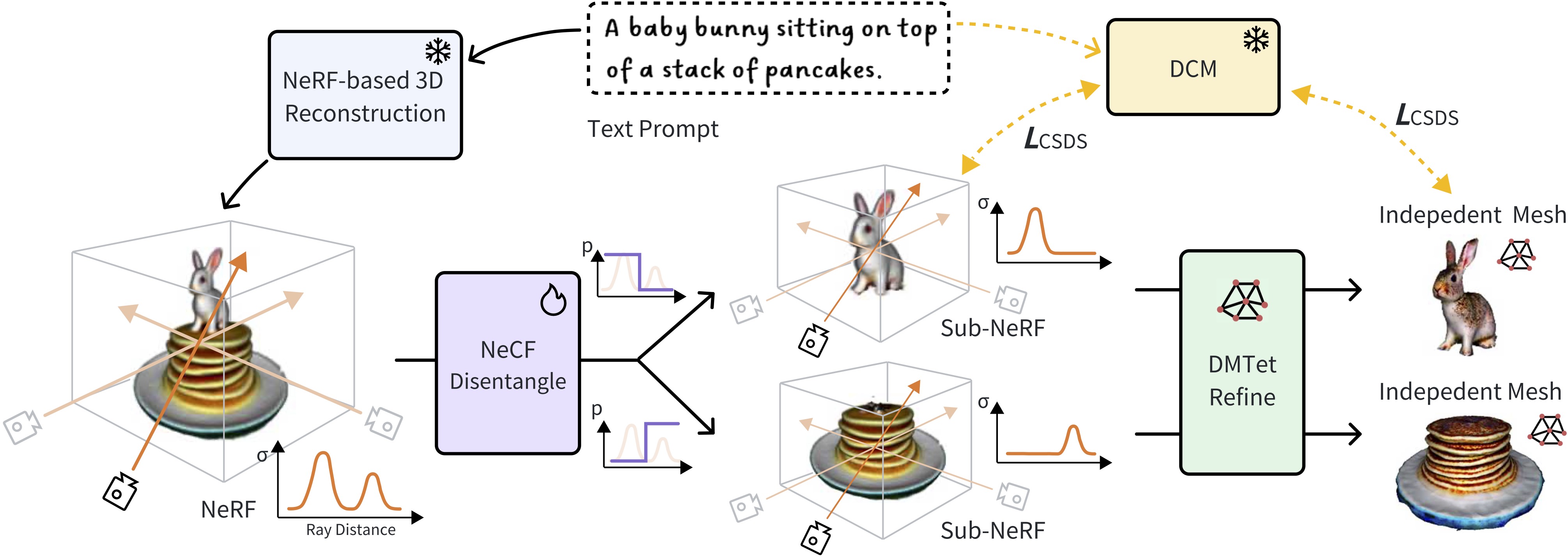}\\
    \caption{\textbf{Method overview.} 
    We generate multiple independent interactive 3D objects in a coarse-to-fine manner. Initially, we render a view of the input text-to-3D NeRF for Deep Concept Mining (DCM), obtaining both the T2I diffusion model and the corresponding text embedding. We then use the mined embedding and the T2I diffusion model to train the neural category field (NeCF) using category score distillation sampling (CSDS). After disentangling the input NeRF, we convert the sub-NeRFs into DMTets and fine-tune these for further refinement. Finally, we export independent surface meshes with improved geometries and textures.
    }
    \label{fig:pipeline}
\end{figure*}
\subsection{Overview}
\hspace{10pt}
DreamDissector begins with a text-to-3D Neural Radiance Field (NeRF). Its objective is to disentangle the generated 3D NeRF into separate 3D assets according to the object categories that the NeRF contains. To achieve this, we introduce a 3D representation called the Neural Category Field~(NeCF). This is designed to disentangle the target NeRF into multiple sub-NeRFs while maintaining the original appearance of each object. The NeCF is supervised by our newly introduced Category Score Distillation Sampling (CSDS), an approach that involves a series of Score Distillation Samplings (SDS) conditioned on category-specific text prompts for the sub-NeRFs. Subsequently, the sub-NeRFs are transformed into DMTets~\cite{dmtet} for final geometry and texture refinement. Since DMTets can be readily converted into surface meshes, DreamDissector ultimately produces independent surface meshes for each object with the preservation of the actions and interactions, thereby facilitating editing by human artists. An outline of our DreamDissector framework is illustrated in Figure~\ref{fig:pipeline}.

\subsection{Neural Category Field}
\hspace{10pt}
To render each categorical object in the target NeRF, a straightforward solution is to introduce a sub-NeRF for each object, \eg, a density field and a color field, respectively. Subsequently, each object can be rendered using its density and color field. The entire NeRF can then be rendered by composing these density and color fields according to the principles of volume rendering~\cite{niemeyer2021giraffe, drebin1988volume}:
\begin{equation}
\sigma = \sum_{k=1}^{K}\sigma_k, \quad \quad \mathbf{c} = \frac{1}{\sigma} \sum_{k=1}^{K}\sigma_k \mathbf{c}_k,
\label{eq:composation}
\end{equation}
where $K$ denotes the number of categories. However, this approach requires training additional networks for density and color fields and needs a constraint loss to maintain the appearance consistency of the entire NeRF.

To this end, we propose an alternate formulation for rendering each category object by \textit{\textbf{de-composing}} the density field with a probability distribution, namely the category field. Specifically, the above density composition can be reformulated as follows,
\begin{align}
\sigma= \sum_{k=1}^{K} \frac{\sigma_k}{\sigma} \sigma, \quad \frac{\sigma_k}{\sigma} \in [0, 1]
\end{align}
Note that a small number can be added to the density to avoid division by zero. 
Consequently, the $\frac{\sigma_k}{\sigma}$ can be regarded as a probability simplex~\cite{boyd2004convex} as it sums to one, and its elements are non-negative. Inspired by this, we leverage an MLP with the softmax function to model the probability simplex $\frac{\sigma_k}{\sigma}$ directly. 
Let $\mathbf{p}_i^k$ represents the probability of the $i$-th point in the 3D space belongs to the $k$-th category:
\begin{equation}
    \mathbf{p}_i^k = \frac{\exp(f_k / T)}{\sum_k^K\exp(f_k /T)}, \quad \mathbf{p}_i^k \in [0, 1]
\end{equation}
where $T$ denotes the temperature, which controls the sharpness of the probability distribution, $f \in \mathbb{R}^K$ is the output of the category field network. With the category field, the color of the $k$-th category object can be rendered as follows,
\begin{equation}
    C(\mathbf{r})^k = \sum_{i=1}^{N} \alpha_i^k(1 - \exp(-\mathbf{p}_i^k\sigma_i\delta_i))\mathbf{c}_i,
     \quad \alpha_i^k = \exp(-\sum_{j=1}^{i-1}\mathbf{p}_i^k\sigma_j\delta_j).
\end{equation}
It can be observed that the density $\sigma$ for each point is scaled by the category field $\mathbf{p}^k$. In other words, we can interpret $\mathbf{p}^k\sigma$ as the density of the sub-NeRF for the $k$-th category object, \eg, $\sum_k\mathbf{p}_i^k = 1$ and $\sum_k(\mathbf{p}_i^k\sigma) = \sigma$. Furthermore, the color field can be reused and frozen during training, simplifying the training process.

Notably, the design of the NeCF has the following merits: (1) We only need to train an additional category field network, which is more efficient than training additional density and color fields networks. 
(2) As the original density and color field networks are frozen during training, the re-composition of the sub-NeRFs exactly equals the original NeRF, preserving its original appearance.

\begin{figure}[t]
\begin{minipage}[htbp]{0.3\linewidth}
\centering
\includegraphics[width=1.0\linewidth]{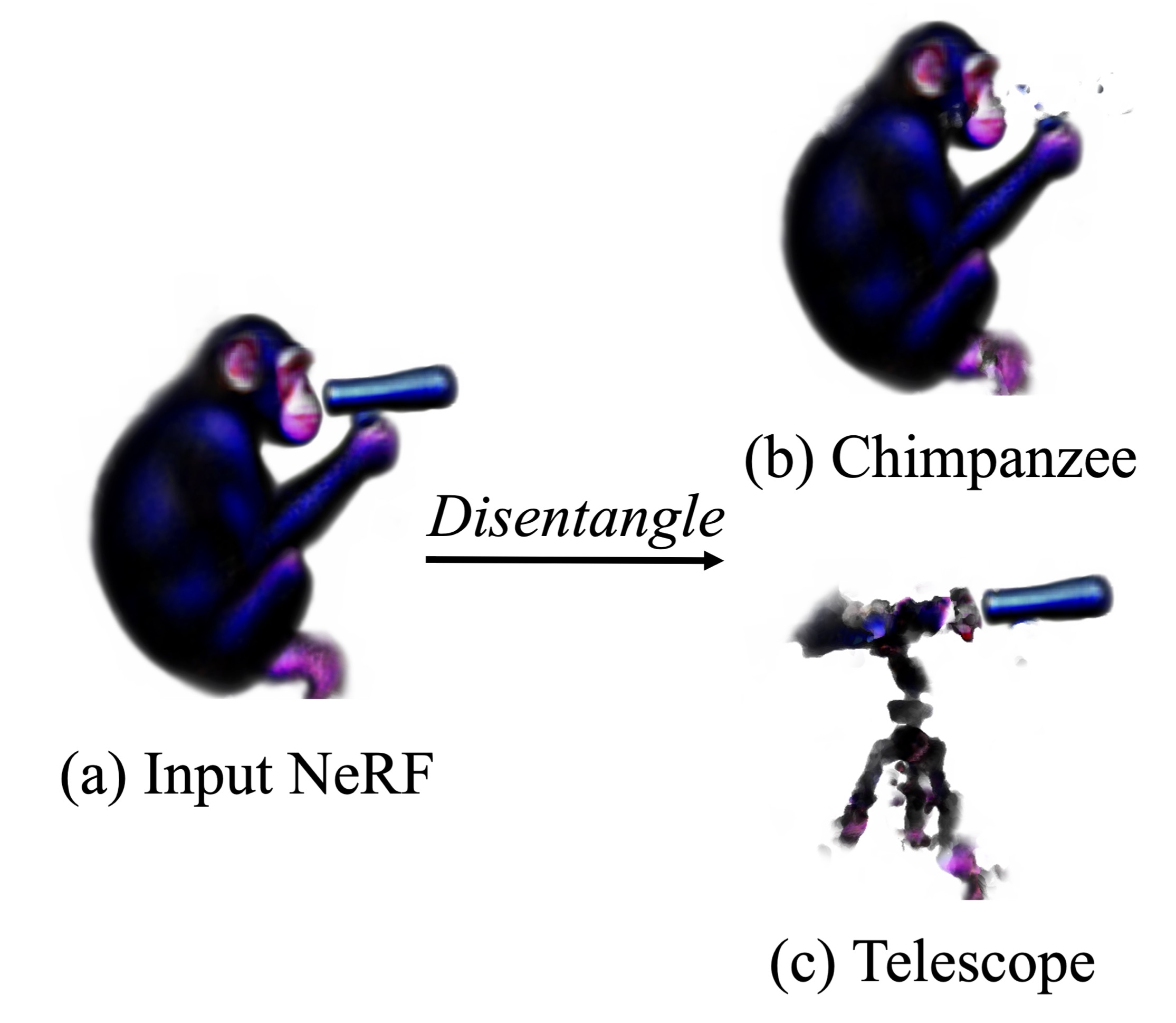}
\end{minipage}
\begin{minipage}[htbp]{0.7\linewidth}
\centering
\includegraphics[width=0.85\linewidth]{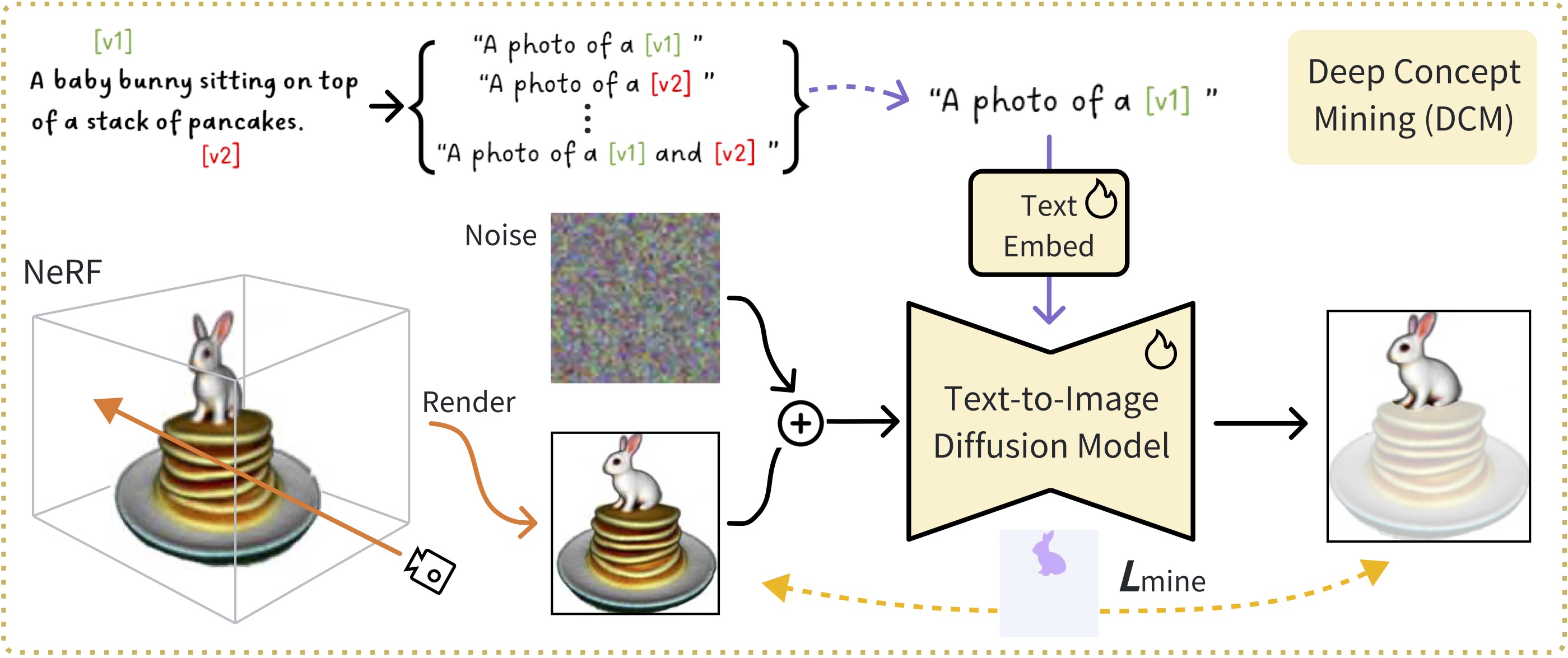}
\end{minipage}
\caption{
\textbf{Left: Concept discrepancy in diffusion models.} The text prompt is \textit{``A chimpanzee looking through a telescope''}.
\textbf{Right: Overview of Deep Concept Mining~(DCM).} We finetune the text embedding and the T2I diffusion model with the masked diffusion loss~(Eq.~\ref{eq:masked_diffusion}).
}
\label{fig:discrepancy_dcm}
\end{figure}

\subsection{Category Score Distillation Sampling}
\label{sec:csds}
\noindent\textbf{A naive approach.}
To train the NeCF, one naive approach is to employ multiple SDS losses to supervise the category field for each category. Specifically, for the object of the $k$-th category, the gradients of its SDS loss can be formulated as,
\begin{equation}
\nabla_\theta L_{SDS}(\phi, \theta)_k = \mathbb{E}_{t, \epsilon} \left[ w(t) \left( \epsilon_\phi(x_t; y_k, t) - \epsilon \right) \frac{\partial x}{\partial \theta} \right],
\end{equation}
where $y_k$ denotes the text embedding for the $k$-th category. For example, given a NeRF generated by the prompt: ``\textit{a} [$v_1$] \textit{sitting on a [$v_2$].}'', the text prompts for the category objects would be ``\textit{a} [$v_1$]'' and ``\textit{a} [$v_2$]''. This can be easily accomplished by human users or by modern LLMs. It is important to note that we do not require SDS for the entire text prompts for training NeCF, and all networks, except for the category field networks, are frozen.

\noindent\textbf{Concept Discrepancy in Diffusion Model.}
Although the naive approach can handle some simple cases, it fails to disentangle scenes with concept gaps from the text descriptions. The concept gap refers to the discrepancy where the object generated by the complete text prompt and by the category text prompt occupy different areas in the 2D diffusion model's latent space. For instance, the text prompt ``\textit{a chimpanzee looking through a telescope.}'' would generate a scene depicting a chimpanzee using a handheld telescope, as shown in Figure~\ref{fig:discrepancy_dcm} left~(a). In contrast, the category text prompt ``\textit{a telescope}'' is more likely to generate a tripod-mounted telescope, since the tripod-mounted telescope is situated in the dominant feature space of the prompt ``\textit{a telescope}'' while the handheld telescope occupies a marginal feature space. As a result, the learned NeCF will produce a tripod-mounted telescope with the tripod being hidden inside the chimpanzee's body, as illustrated by Figure~\ref{fig:discrepancy_dcm} left~(c).

\noindent\textbf{Deep Concept Mining.}
To address this issue, we propose mining the concepts in the text prompt and aligning them with ones depicted in NeRF for disentanglement, as illustrated in Figure~\ref{fig:discrepancy_dcm} right. To this end, a T2I diffusion model is personalized to denoise a given view rendered by the NeRF into an image depicting one (or several) independent object(s), under the condition of one (or several) specific concept(s).
Specifically, we first create a set of prompts where each contains one or several concepts. For each concept or concept combination, we generate the corresponding segmentation mask for a rendered view of the NeRF, through a text-based open-vocabulary segmentation model, \eg Grounded-SAM~\cite{ren2024grounded}. Then we utilize the prompt-mask pairs to optimize the text embedding and the diffusion backbone with a concept mining loss with mask attention~\cite{avrahami2023break}:
\begin{equation}
L_{mine}(\phi, y_k) = \mathbb{E}_{t, \epsilon} \left[ || \epsilon_\phi(x_t; y_k, t) \odot M_k- \epsilon \odot M_k ||_2^2 \right],
\label{eq:masked_diffusion}
\end{equation}
where $M_k$ denotes the mask of the $k$-th category. 
The DCM module is frozen after optimization to provide better alignment between the independent textual concepts and the sub-NeRFs for better NeCF training with the CSDS loss. The frozen DCM is also used to train the DMTet refinement module as illustrated in Figure~\ref{fig:pipeline}.

\noindent\textbf{Final Refinement.}
After training NeCF, we convert the sub-NeRFs into DMTets using the isosurface extraction technique~\cite{lin2023magic3d}, and fine-tune these DMTets with text embeddings and the model from DCM. 
The rationale is that further refinement can fix the artifacts produced by disentanglement, and DMTets can be easily converted into surface meshes.
However, DCM tends to overfit the mined concept in the original NeRF, resulting in over-saturated and unrealistic colors~\cite{poole2022dreamfusion,avrahami2023break}. To address this, we employ the original stable diffusion to fine-tune the colors of the DMTets through additional steps, enhancing their realism. Finally, the DMTets are transformed into textured meshes.

\begin{figure*}[t]
    \centering
    \includegraphics[width=\linewidth]{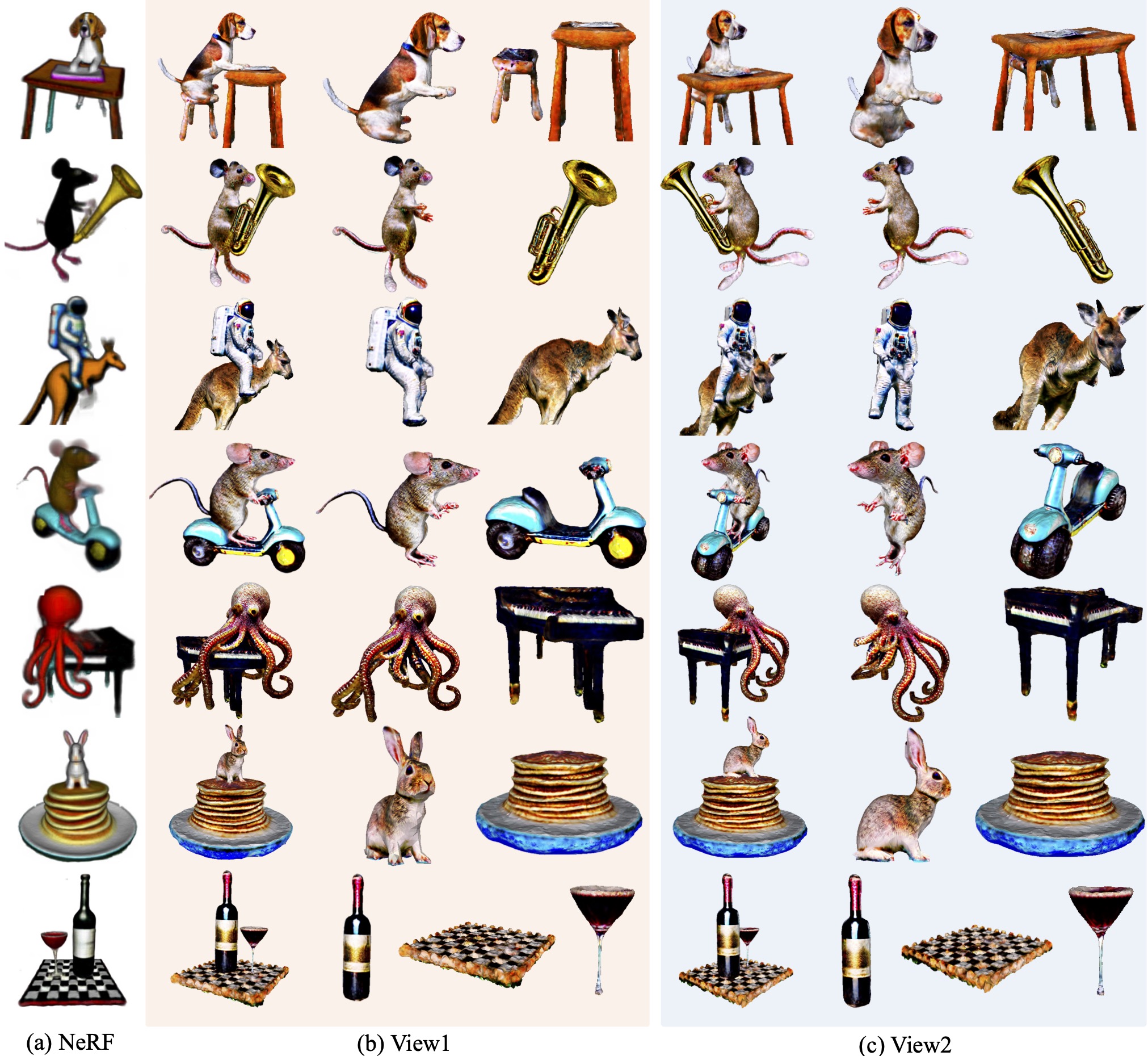}\\
    \caption{\textbf{Qualitative Results.} The text prompts used to generate the NeRF are available in the supplementary file.
    }
    \label{fig:results_7_cases}
\end{figure*}
\subsection{Overall Pipeline}
Therefore, the overall pipeline comprises the following steps:
\begin{itemize}
    \item Generate a mask for each category in the prompt from a rendered view and use them to optimize the DCM module.
    \item Freeze the DCM, and train the NeCF network using the CSDS loss for decompose a NeRF into sub-NeRFs of independent objects.
    \item Convert the sub-NeRFs into DMTets and fine-tune them with the optimized DCM module, then fine-tune the colors of the DMTets using the original stable diffusion to produce the final output.
\end{itemize}

\section{Experiments}
\subsection{Implementation Details}
\label{sec:implement}
\textbf{Category field.}
Our method is implemented based on the publicly available repository threestudio\footnote{https://github.com/threestudio-project/threestudio}. We use instant-NGP\cite{muller2022instantngp} as the NeRF representation, which includes a multi-resolution hash grid and two MLPs for the density and color networks. Similarly, we employ a random initialized hash grid and an MLP for the category field, using the same configuration as the density and color networks. The output dimension of the category field MLP is set to the number of categories in the original NeRF, and the temperature $T$ is set to 0.05 to make the category field approximate a one-hot encoding. The training of the category field follows the Dreamfusion approach, except that the color and density networks are frozen. The training lasts for 1,000 steps with a batch size of 1, taking approximately 3 minutes.

\noindent \textbf{Deep concept mining.}
For DCM, we have adopted Stable Diffusion 2.1 as the backbone. Inspired by~\cite{avrahami2023break}, we adopt a two-stage training strategy for deep concept mining to stabilize the training: fine-tuning the text embedding in the first stage and fine-tuning both the text embedding and the backbone in the second stage. We use a learning rate of $5 \times 10^{-4}$ for the first stage and a learning rate of $2 \times 10^{-6}$ for the second stage. The first stage is trained for 400 steps, and the second stage for 100 steps. Notably, the training of DCM takes approximately 6 minutes on an NVIDIA A100, which is time-efficient compared with most text-to-3D methods that require several hours of training on a single GPU, including Dreamfusion~\cite{poole2022dreamfusion}, Magic3D~\cite{lin2023magic3d}, ProlificDreamer~\cite{wang2023prolificdreamer}, and MVDream~\cite{shi2023mvdream}.

\noindent \textbf{Final refinement.}
We adopt the same technique used in Magic3D~\cite{lin2023magic3d} to convert the sub-NeRFs into DMTets. Similarly, we train the DMTets with a batch size of 8 for 5,000 steps. Subsequently, we fix the DMTet geometry and fine-tune the color for 1,000 steps using the original Stable Diffusion. Note that the color fine-tuning is optional as the color is already decent for most cases. 
During color fine-tuning, we use ``unrealistic, low quality, shadow'' as the negative prompt to improve quality.
Finally, we export the DMTets as textured meshes. 

\subsection{Results}
\label{sec:results}
\noindent \textbf{Main results.}
\begin{figure*}[t]
    \centering
    \includegraphics[width=1\linewidth]{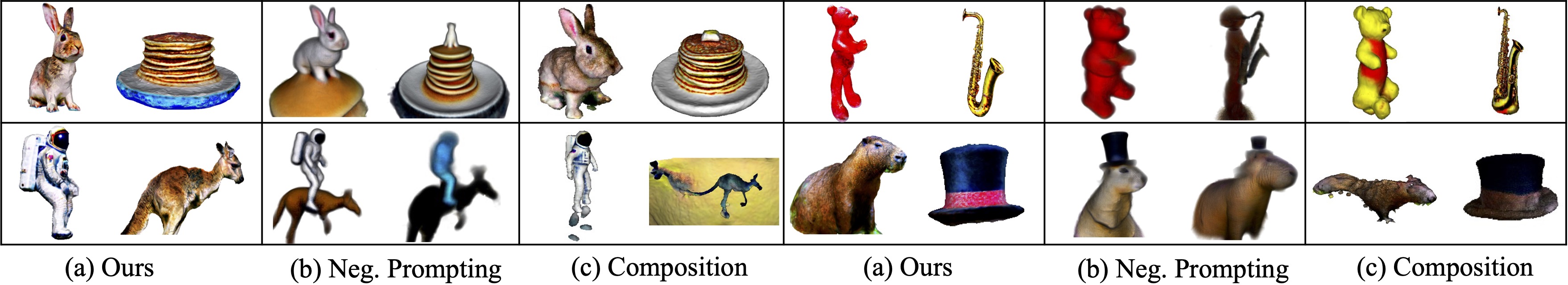}\\
    \caption{\textbf{Comparison with two baselines.} We show the independent objects for ease of comparison. 
    The composed objects and more comparisons are available in the supplementary file.
    }
    \label{fig:comparsion}
\end{figure*}
\begin{figure*}[th!]
    \centering
    \includegraphics[width=\linewidth]{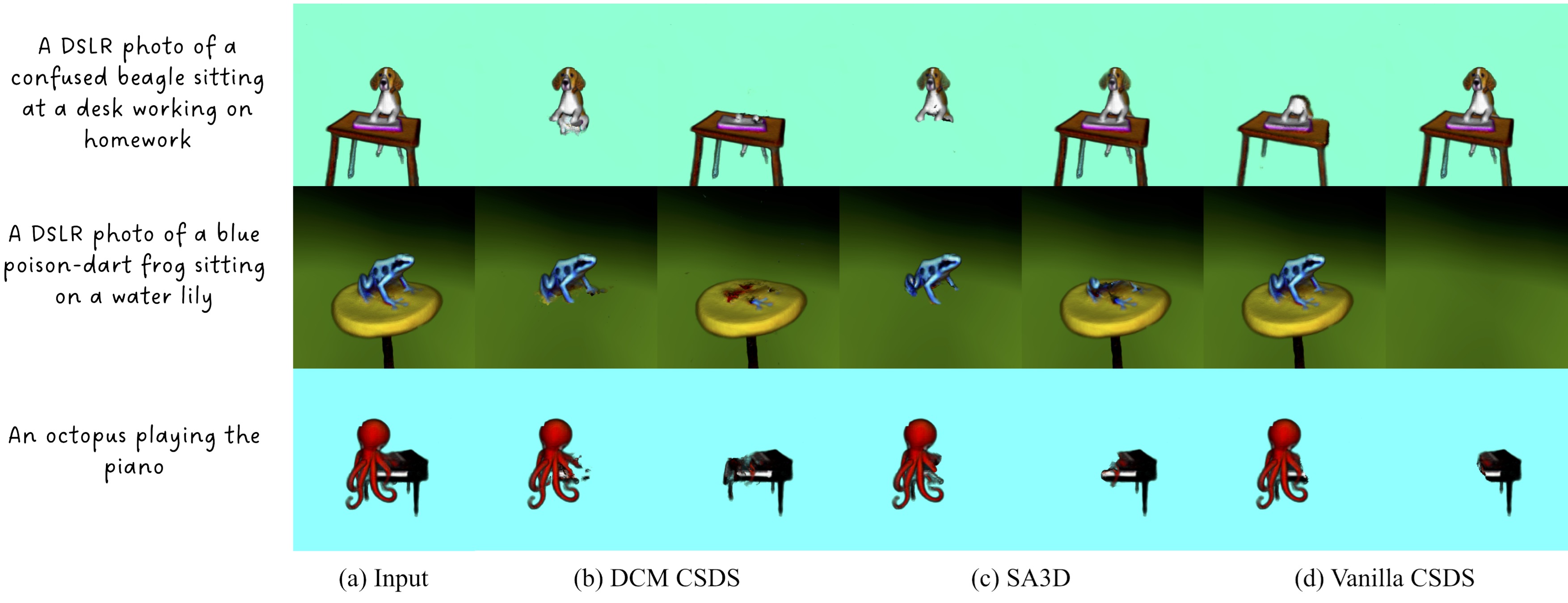}\\
    \caption{
    \textbf{Using different strategies to disentangle the NeRF.} 
    Our DCM CSDS successfully disentangles the sub-NeRFs, whereas SA3D~\cite{cen2023sa3d} and the vanilla CSDS fail in some cases. Note that the outliers can be easily removed by thresholding in the isosurface extraction.}
    \label{fig:ablation1}
\end{figure*}
The qualitative results are shown in Figure~\ref{fig:results_7_cases}. For each case, two views of each object are sampled, and the corresponding text prompts are available in the supplementary file. It can be observed that DreamDissector can effectively disentangle input scenes featuring various complex interactions, such as riding. Notably, DreamDissector can handle cases with large and complex contact surfaces, as demonstrated in the ``\textit{an octopus playing the piano}'' case, where the octopus's tentacles are disentangled from the piano. Additionally, the final meshes exhibit more realistic and higher-quality textures compared to those in the input NeRF. This improvement is attributed to the final refinement, further demonstrating DreamDissector's practicality.

\noindent \textbf{Comparison.}
\begin{table}[t]
    \centering
    \caption{\textbf{CLIP Score comparisons.}}
    \setlength\tabcolsep{10pt}
    \scalebox{0.8}{
    \begin{tabular}{c|ccc}
    \toprule
    \textbf{Method} & CLIP-B-16 & CLIP-B-32 & CLIP-L-14 \\
    \midrule
    Negative Prompting &  0.299 & 0.296  & 0.247 \\
    Composition  & 0.281  & 0.278  & 0.234 \\
    Ours  & \textbf{0.316} & \textbf{0.311}  & \textbf{0.270} \\
    \bottomrule
    \end{tabular}
    }
    \label{table:rprecision}
\end{table}
We compare DreamDissector with two baselines: negative prompting and a composition baseline. Negative prompting involves taking the entire text prompt as the positive prompt and identifying the exclusive object as the negative prompt. For example, in the prompt ``\textit{a} [$v_1$] \textit{sitting on} [$v_2$]'', the positive prompt for both objects is the entire prompt, while the negative prompt for object [$v_1$] is ``[$v_2$]'', and vice versa.
Since the most relevant works, CompoNeRF~\cite{lin2023componerf} and Comp3D~\cite{po2023compositional}, are not open-source, we implemented a composition baseline with a similar idea: training the objects separately and then composing them with further fine-tuning. We compare our method with these baselines both qualitatively and quantitatively. As Figure~\ref{fig:comparsion} demonstrates, DreamDissector significantly outperforms the baseline methods.
Additionally, we evaluate DreamDissector and the baseline methods using the CLIP Score metric, which measures cosine similarity between the embeddings of the text and image. We conduct this evaluation on both independent and composed objects and compute the average score. As Table~\ref{table:rprecision} shows, our method significantly outperforms the baselines.

\subsection{Analysis}
\noindent \textbf{DCM for disentanglement.}
To evaluate the effectiveness of deep concept mining (DCM) in disentangling NeRF, we conducted a comparative analysis with two baselines: (1) the vanilla Category Score Distillation Sampling (CSDS) without DCM, and (2) Segment Anything~\cite{kirillov2023sam} in 3D with NeRFs (SA3D)~\cite{cen2023sa3d}. Unlike the fully unsupervised vanilla CSDS, both our DCM approach and SA3D require an input mask for a single view. As Figure~\ref{fig:ablation1} shows, the vanilla CSDS struggles to disentangle NeRF for scenarios with significant concept discrepancies, such as ``\textit{a blue poison-dart frog sitting on a water lily},'' where the original scene primarily depicts water lily leaves. While SA3D manages to disentangle scenarios involving concept discrepancies, such as the frog, it falls short in more complex cases with extensive occlusions, exemplified by the beagle and octopus cases. In contrast, DCM demonstrates superior performance, successfully disentangling scenarios involving concept discrepancies and significant occlusions. 

\begin{figure}[t]
    \centering
    \includegraphics[width=0.9\linewidth]{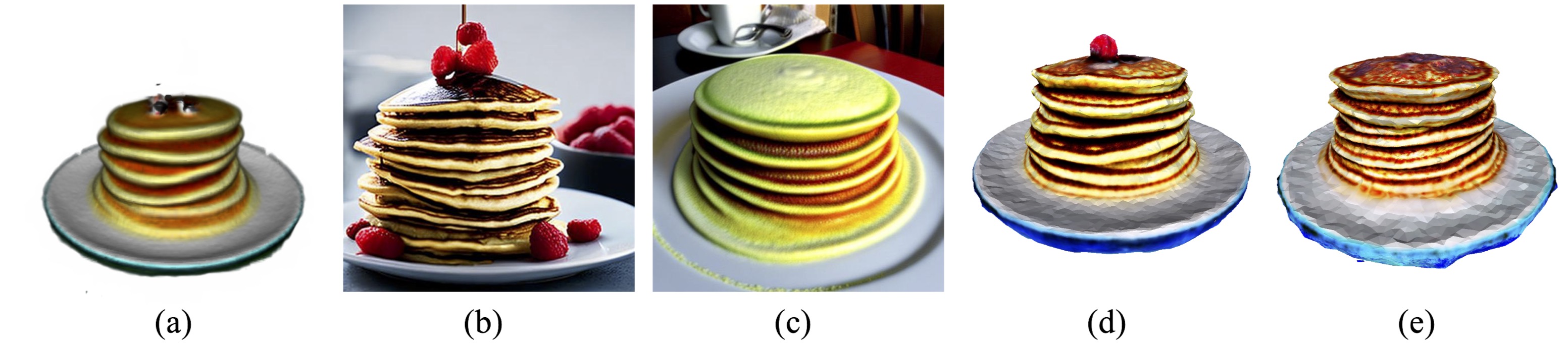}\\
    \caption{
    \textbf{Illustration of DCM refinement.}
    (a) Disentangled NeRF with artifacts, 
    (b) sampled image of original stable diffusion~(SD),
    (c) sampled image of DCM SD,
    (d) undesired results from fine-tuning on original SD, and 
    (e) artifacts fixed by DCM SD.
    }
    \label{fig:refinement}
\end{figure}
\noindent \textbf{DCM for refinement.}
DCM is utilized not only for NeRF disentanglement but also for refining DMTets. We conducted an analysis on the effectiveness of DCM in this refinement. The results are presented in Figure~\ref{fig:refinement}. From (a), it can be observed that artifacts remain after disentanglement. Due to the original NeRF's invisible contact surface, a ``black hole'' appears post-disentanglement. However, using original stable diffusion for DMTet refinement doesn't resolve this, as illustrated in (d). This is because the prompt ``\textit{a stack of pancakes}'' typically generates images with fruits on the pancakes, as these are prevalent in the high-density regions of stable diffusion, evidenced in (b). Consequently, the fine-tuned DMTet produces fruit from the black hole artifact region. In contrast, DCM stable diffusion closely matches the input pancake, as shown in the first row, effectively fixing the artifacts during DMTet refinement, as seen in (e). This further demonstrates DCM's superiority.

\begin{figure}[t]
    \centering
    \includegraphics[width=0.9\linewidth]{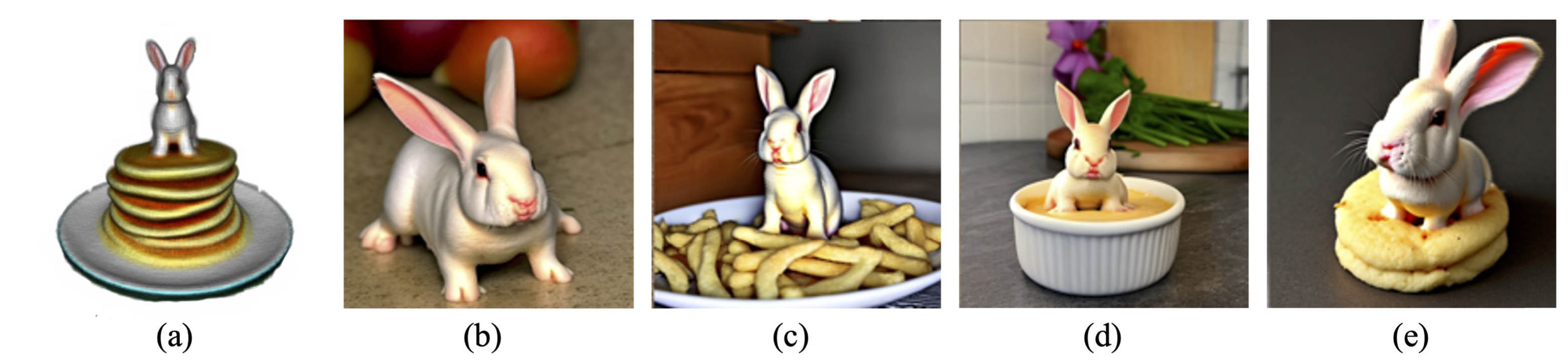}\\
    \caption{
    \textbf{Ablation study of DCM.}
    We show the sampled images of DCM fine-tuned model.
    (a) The input rendered image,
    (b) DCM,
    (c) without the masked attention loss,
    (d) without the first stage training, and
    (e) without the second stage training.
    }
    \label{fig:dcm_ablation}
\end{figure}
\noindent \textbf{Ablation study on DCM.} 
We perform an ablation study on each component of DCM, including two-stage training and the masked attention loss. Specifically, we sample the images of the mined concept of ``\textit{baby bunny}'' from the text prompt ``\textit{a baby bunny sitting on a stack of pancakes}'', using the fine-tuned models. Ideally, the sampled image should not contain any concepts similar to pancakes.
As Figure~\ref{fig:dcm_ablation} shows, DCM successfully extracts the concept of ``\textit{baby bunny}'', whereas other training strategies fail to separate this concept from others, such as items resembling the pancakes upon which it is sitting. This demonstrates DCM's ability to mine independent concepts.

\subsection{Applications}
\begin{figure}[t]
    \centering
    \includegraphics[width=0.99\linewidth]{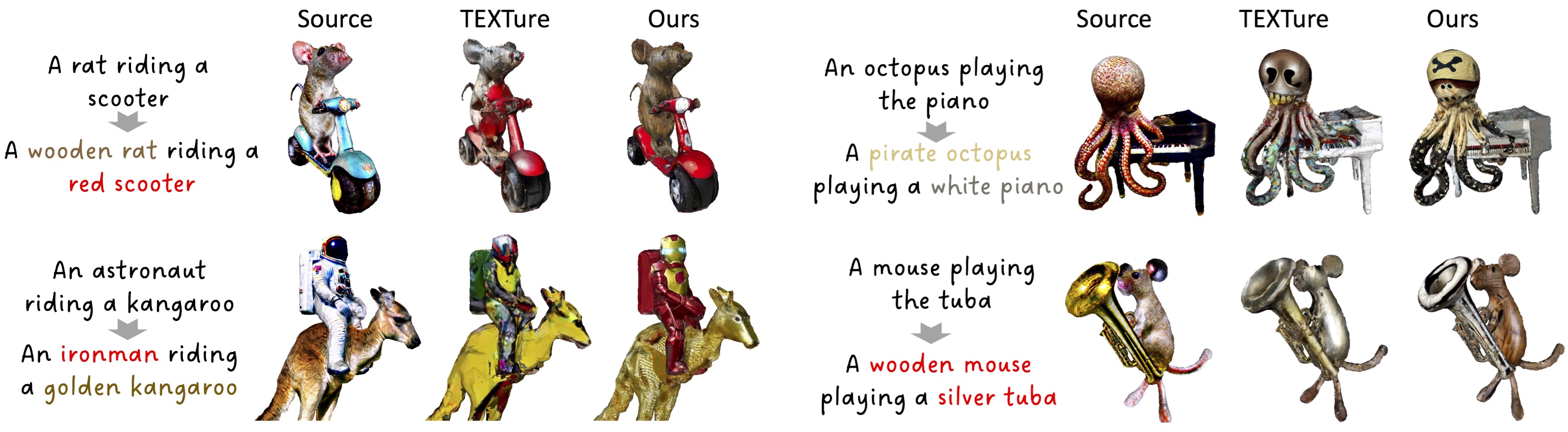}\\
    \caption{
    Illustration of the application of \textbf{text-guided texture editing}.
    }
    \label{fig:texture_application}
\end{figure}

\noindent \textbf{Controllable texture editing.}
Although text-guided texturing has achieved notable progress~\cite{chen2023text2tex,richardson2023texture}, generating textures for complex scenes with multiple objects remains challenging. We evaluated TEXTure~\cite{richardson2023texture} in three different cases, as depicted in Figure~\ref{fig:texture_application}. For the baseline, We treated the multi-object mesh as a single entity and applied TEXTure. For our approach, we applied TEXTure to each object's mesh individually and then combined them.
We observed that the textures generated by the baseline approach poorly matched the input prompts and were of low quality. Notably, textures for independent objects were influenced by other objects in the scene, \eg, part of the rat exhibited a red color. In contrast, DreamDissector significantly enhances TEXTure's performance, producing visually appealing and accurate textures.

\begin{figure}[t]
    \centering
    \includegraphics[width=0.99\linewidth]{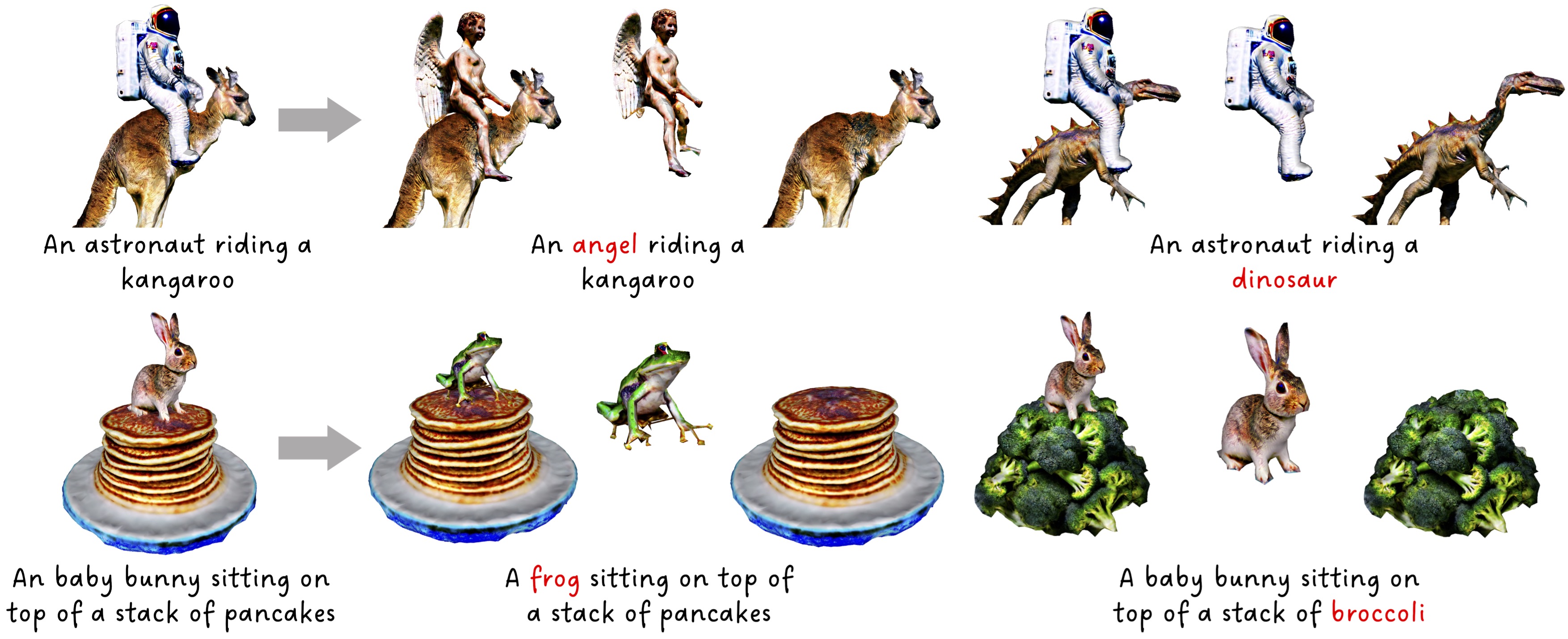}\\
    \caption{
    Illustration of the application of \textbf{text-guided object replacement}.
    }
    \label{fig:concept_application}
\end{figure}

\noindent \textbf{Controllable object replacement.}
In addition to controllable texture editing, DreamDissector also has the capability to replace individual objects without affecting other objects in the scene. To achieve this, the target DMTet is fine-tuned while the remaining DMTets are kept fixed. However, deforming a DMTet into a completely different topology object is challenging using SDS-based supervision~\cite{lin2023magic3d}. Inspired by ~\cite{chen2023fantasia3d}, we initially feed the normals of the DMTet to stable diffusion for several steps, effectively deforming the DMTet. 
We also empirically observe that fine-tuning the target DMTet only will induce severe mesh interpenetration. To address this, we introduce an interpenetration loss,
\begin{equation}
\mathcal{L}_{interpenetration} = \sum_{i} \max(\epsilon - (\mathbf{v}_i - \mathbf{v}_i') \cdot \mathbf{n}_i', 0),
\end{equation}
where $\mathbf{v}_i$ represents the $i$-th vertex of the target DMTet, $\mathbf{v}_i'$ and $\mathbf{n}_i'$ are the vertex and vertex normal of the nearest neighbor of $\mathbf{v}_i$ in other DMTets, respectively, and $\epsilon$ is a small tolerance hyper-parameter for interpenetration. The results, shown in Figure~\ref{fig:concept_application}, demonstrate that DreamDissector can achieve controllable concept replacement.

\begin{figure}[t]
    \centering
     \includegraphics[width=0.8\linewidth]{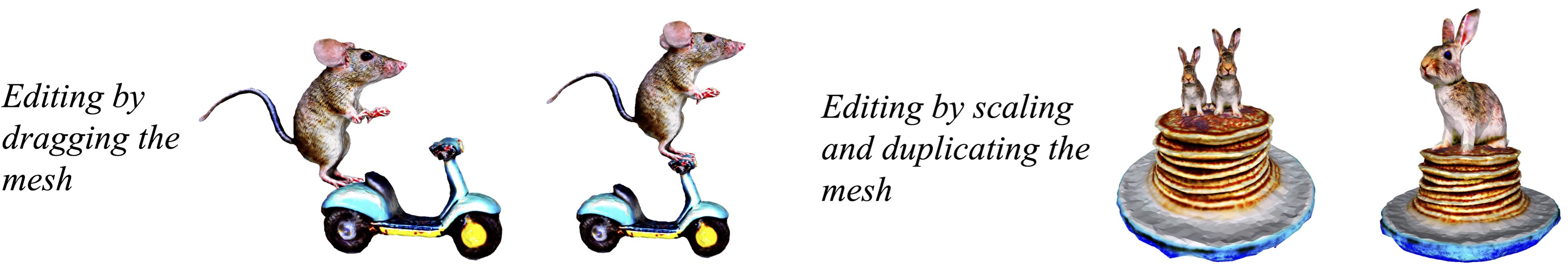}\\
    \caption{
    Illustration of the application of \textbf{geometry editing}.
    }
    \label{fig:application_geo}
\end{figure}
\noindent \textbf{Geometric editing by users.}
To further verify how DreamDissector can facilitate user workflows, we allowed a user to edit objects individually. As demonstrated in Figure~\ref{fig:application_geo}, an object can be easily modified by simple operations such as scaling, translation, and dragging, thereby highlighting DreamDissector's effectiveness in enhancing human editing capabilities in practical applications.
\section{Conclusion}
\hspace{10pt}
We propose DreamDissector, a novel framework designed to generate multiple, independently interacting objects guided by text. DreamDissector takes a multi-object text-to-3D Neural Radiance Field (NeRF) as input and produces several textured meshes. We introduce the Neural Category Field (NeCF), a representation capable of disentangling the input NeRF into multiple sub-NeRFs. To train NeCF, we present the Category Score Distillation Sampling (CSDS) loss. Moreover, we have observed a concept discrepancy issue in 2D diffusion models, which can degrade disentanglement performance. To address this, we introduce Deep Concept Mining (DCM) to fine-tune the text embeddings and the 2D diffusion model, effectively deriving sub-NeRFs. Additionally, we propose a two-stage refinement process to further refine the geometry and texture, thereby enhancing realism.
Experimental results and further applications showcase the effectiveness and practicality of DreamDissector in real-world scenarios.

\noindent\textbf{Acknowledgement}
The work was supported in part by NSFC with Grant No.~62293482, 
the Basic Research Project No HZQB-KCZYZ-2021067 of Hetao Shenzhen-HK S\&T Cooperation Zone, Guangdong Provincial Outstanding Youth Fund (No.~2023B1515020055), the National Key R\&D Program of China with grant No.~2018YFB1800800, by Shenzhen Outstanding Talents Training Fund 202002, by Guangdong Research Projects No.~2017ZT07X152 and No.~2019CX01X104, by Key Area R\&D Program of Guangdong Province (Grant No. 2018B030338001), by the Guangdong Provincial Key Laboratory of Future Networks of Intelligence (Grant No. 2022B1212010001), and by Shenzhen Key Laboratory of Big Data and Artificial Intelligence (Grant No. ZDSYS201707251409055). It is also partly supported by NSFC-61931024, NSFC-62172348, and Shenzhen Science and Technology Program No. JCYJ20220530143604010.




%
%
\bibliographystyle{splncs04}
\bibliography{main}

\begin{thebibliography}{10}
\providecommand{\url}[1]{\texttt{#1}}
\providecommand{\urlprefix}{URL }
\providecommand{\doi}[1]{https://doi.org/#1}

\bibitem{achlioptas2018learning}
Achlioptas, P., Diamanti, O., Mitliagkas, I., Guibas, L.: Learning representations and generative models for 3d point clouds. In: ICML (2018)

\bibitem{avrahami2023break}
Avrahami, O., Aberman, K., Fried, O., Cohen-Or, D., Lischinski, D.: Break-a-scene: Extracting multiple concepts from a single image. arXiv preprint arXiv:2305.16311  (2023)

\bibitem{bar2022text2live}
Bar-Tal, O., Ofri-Amar, D., Fridman, R., Kasten, Y., Dekel, T.: Text2live: Text-driven layered image and video editing. In: ECCV (2022)

\bibitem{boyd2004convex}
Boyd, S.P., Vandenberghe, L.: Convex optimization. Cambridge university press (2004)

\bibitem{cen2023sa3d}
Cen, J., Zhou, Z., Fang, J., Shen, W., Xie, L., Zhang, X., Tian, Q.: Segment anything in 3d with nerfs. arXiv preprint arXiv:2304.12308  (2023)

\bibitem{chen2023text2tex}
Chen, D.Z., Siddiqui, Y., Lee, H.Y., Tulyakov, S., Nie{\ss}ner, M.: Text2tex: Text-driven texture synthesis via diffusion models. arXiv preprint arXiv:2303.11396  (2023)

\bibitem{chen2023fantasia3d}
Chen, R., Chen, Y., Jiao, N., Jia, K.: Fantasia3d: Disentangling geometry and appearance for high-quality text-to-3d content creation (2023)

\bibitem{chen2019learning}
Chen, Z., Zhang, H.: Learning implicit fields for generative shape modeling. In: CVPR (2019)

\bibitem{set-the-scene}
Cohen-Bar, D., Richardson, E., Metzer, G., Giryes, R., Cohen-Or, D.: Set-the-scene: Global-local training for generating controllable nerf scenes. In: ICCVW (2023)

\bibitem{drebin1988volume}
Drebin, R.A., Carpenter, L., Hanrahan, P.: Volume rendering. Siggraph  (1988)

\bibitem{epstein2024disentangled}
Epstein, D., Poole, B., Mildenhall, B., Efros, A.A., Holynski, A.: Disentangled 3d scene generation with layout learning. arXiv preprint arXiv:2402.16936  (2024)

\bibitem{fridman2023scenescape}
Fridman, R., Abecasis, A., Kasten, Y., Dekel, T.: Scenescape: Text-driven consistent scene generation (2023)

\bibitem{gao2024graphdreamer}
Gao, G., Liu, W., Chen, A., Geiger, A., Sch{\"o}lkopf, B.: Graphdreamer: Compositional 3d scene synthesis from scene graphs. In: CVPR (2024)

\bibitem{gao2022get3d}
Gao, J., Shen, T., Wang, Z., Chen, W., Yin, K., Li, D., Litany, O., Gojcic, Z., Fidler, S.: Get3d: A generative model of high quality 3d textured shapes learned from images. NeurIPS  (2022)

\bibitem{henzler2019escaping}
Henzler, P., Mitra, N.J., Ritschel, T.: Escaping plato's cave: 3d shape from adversarial rendering. In: ICCV (2019)

\bibitem{ho2020denoising}
Ho, J., Jain, A., Abbeel, P.: Denoising diffusion probabilistic models. NeurIPS  (2020)

\bibitem{hong2022avatarclip}
Hong, F., Zhang, M., Pan, L., Cai, Z., Yang, L., Liu, Z.: Avatarclip: Zero-shot text-driven generation and animation of 3d avatars. Siggraph  (2022)

\bibitem{hong2023debiasing}
Hong, S., Ahn, D., Kim, S.: Debiasing scores and prompts of 2d diffusion for view-consistent text-to-3d generation (2023)

\bibitem{hong20233dllm}
Hong, Y., Zhen, H., Chen, P., Zheng, S., Du, Y., Chen, Z., Gan, C.: 3d-llm: Injecting the 3d world into large language models (2023)

\bibitem{huang2023dreamtime}
Huang, Y., Wang, J., Shi, Y., Qi, X., Zha, Z.J., Zhang, L.: Dreamtime: An improved optimization strategy for text-to-3d content creation (2023)

\bibitem{höllein2023text2room}
Höllein, L., Cao, A., Owens, A., Johnson, J., Nießner, M.: Text2room: Extracting textured 3d meshes from 2d text-to-image models (2023)

\bibitem{jain2022zeroshot}
Jain, A., Mildenhall, B., Barron, J.T., Abbeel, P., Poole, B.: Zero-shot text-guided object generation with dream fields. CVPR  (2022)

\bibitem{Mohammad_Khalid_2022}
Khalid, N.M., Xie, T., Belilovsky, E., Popa, T.: {CLIP}-mesh: Generating textured meshes from text using pretrained image-text models. In: Siggraph Asia (2022)

\bibitem{kirillov2023sam}
Kirillov, A., Mintun, E., Ravi, N., Mao, H., Rolland, C., Gustafson, L., Xiao, T., Whitehead, S., Berg, A.C., Lo, W.Y., et~al.: Segment anything. ICCV  (2023)

\bibitem{lee2022understanding}
Lee, H.H., Chang, A.X.: Understanding pure clip guidance for voxel grid nerf models (2022)

\bibitem{lin2023magic3d}
Lin, C.H., Gao, J., Tang, L., Takikawa, T., Zeng, X., Huang, X., Kreis, K., Fidler, S., Liu, M.Y., Lin, T.Y.: Magic3d: High-resolution text-to-3d content creation. CVPR  (2023)

\bibitem{lin2023componerf}
Lin, Y., Bai, H., Li, S., Lu, H., Lin, X., Xiong, H., Wang, L.: Componerf: Text-guided multi-object compositional nerf with editable 3d scene layout. arXiv preprint arXiv:2303.13843  (2023)

\bibitem{liu2023zero1to3}
Liu, R., Wu, R., Hoorick, B.V., Tokmakov, P., Zakharov, S., Vondrick, C.: Zero-1-to-3: Zero-shot one image to 3d object. ICCV  (2023)

\bibitem{liu2023syncdreamer}
Liu, Y., Lin, C., Zeng, Z., Long, X., Liu, L., Komura, T., Wang, W.: Syncdreamer: Generating multiview-consistent images from a single-view image (2023)

\bibitem{lunz2020inverse}
Lunz, S., Li, Y., Fitzgibbon, A., Kushman, N.: Inverse graphics gan: Learning to generate 3d shapes from unstructured 2d data. arXiv preprint arXiv:2002.12674  (2020)

\bibitem{luo2021diffusion}
Luo, S., Hu, W.: Diffusion probabilistic models for 3d point cloud generation. In: CVPR (2021)

\bibitem{melaskyriazi2023realfusion}
Melas-Kyriazi, L., Rupprecht, C., Laina, I., Vedaldi, A.: Realfusion: 360$\deg$ reconstruction of any object from a single image. CVPR  (2023)

\bibitem{mescheder2019occupancy}
Mescheder, L., Oechsle, M., Niemeyer, M., Nowozin, S., Geiger, A.: Occupancy networks: Learning 3d reconstruction in function space. In: CVPR (2019)

\bibitem{metzer2022latentnerf}
Metzer, G., Richardson, E., Patashnik, O., Giryes, R., Cohen-Or, D.: Latent-nerf for shape-guided generation of 3d shapes and textures. CVPR  (2023)

\bibitem{michel2021text2mesh}
Michel, O., Bar-On, R., Liu, R., Benaim, S., Hanocka, R.: Text2mesh: Text-driven neural stylization for meshes. CVPR  (2022)

\bibitem{mildenhall2021nerf}
Mildenhall, B., Srinivasan, P.P., Tancik, M., Barron, J.T., Ramamoorthi, R., Ng, R.: Nerf: Representing scenes as neural radiance fields for view synthesis. Communications of the ACM  (2021)

\bibitem{mo2019structurenet}
Mo, K., Guerrero, P., Yi, L., Su, H., Wonka, P., Mitra, N., Guibas, L.J.: Structurenet: Hierarchical graph networks for 3d shape generation. arXiv preprint arXiv:1908.00575  (2019)

\bibitem{muller2022instantngp}
M{\"u}ller, T., Evans, A., Schied, C., Keller, A.: Instant neural graphics primitives with a multiresolution hash encoding. TOG  (2022)

\bibitem{nichol2022point}
Nichol, A., Jun, H., Dhariwal, P., Mishkin, P., Chen, M.: Point-e: A system for generating 3d point clouds from complex prompts. arXiv preprint arXiv:2212.08751  (2022)

\bibitem{niemeyer2021giraffe}
Niemeyer, M., Geiger, A.: Giraffe: Representing scenes as compositional generative neural feature fields. In: CVPR (2021)

\bibitem{po2023compositional}
Po, R., Wetzstein, G.: Compositional 3d scene generation using locally conditioned diffusion. arXiv preprint arXiv:2303.12218  (2023)

\bibitem{poole2022dreamfusion}
Poole, B., Jain, A., Barron, J.T., Mildenhall, B.: Dreamfusion: Text-to-3d using 2d diffusion. arXiv preprint arXiv:2209.14988  (2022)

\bibitem{radford2021learning}
Radford, A., Kim, J.W., Hallacy, C., Ramesh, A., Goh, G., Agarwal, S., Sastry, G., Askell, A., Mishkin, P., Clark, J., Krueger, G., Sutskever, I.: Learning transferable visual models from natural language supervision. ICML  (2021)

\bibitem{raj2023dreambooth3d}
Raj, A., Kaza, S., Poole, B., Niemeyer, M., Ruiz, N., Mildenhall, B., Zada, S., Aberman, K., Rubinstein, M., Barron, J., Li, Y., Jampani, V.: Dreambooth3d: Subject-driven text-to-3d generation (2023)

\bibitem{ramesh2022hierarchical}
Ramesh, A., Dhariwal, P., Nichol, A., Chu, C., Chen, M.: Hierarchical text-conditional image generation with clip latents. 3DVAR  (2022)

\bibitem{ren2024grounded}
Ren, T., Liu, S., Zeng, A., Lin, J., Li, K., Cao, H., Chen, J., Huang, X., Chen, Y., Yan, F., et~al.: Grounded sam: Assembling open-world models for diverse visual tasks. arXiv preprint arXiv:2401.14159  (2024)

\bibitem{richardson2023texture}
Richardson, E., Metzer, G., Alaluf, Y., Giryes, R., Cohen-Or, D.: Texture: Text-guided texturing of 3d shapes. arXiv preprint arXiv:2302.01721  (2023)

\bibitem{rombach2022high}
Rombach, R., Blattmann, A., Lorenz, D., Esser, P., Ommer, B.: High-resolution image synthesis with latent diffusion models. In: CVPR (2022)

\bibitem{imagen}
Saharia, C., Chan, W., Saxena, S., Li, L., Whang, J., Denton, E.L., Ghasemipour, K., Gontijo~Lopes, R., Karagol~Ayan, B., Salimans, T., et~al.: Photorealistic text-to-image diffusion models with deep language understanding. NeurIPS  (2022)

\bibitem{schuhmann2022laion}
Schuhmann, C., Beaumont, R., Vencu, R., Gordon, C., Wightman, R., Cherti, M., Coombes, T., Katta, A., Mullis, C., Wortsman, M., et~al.: Laion-5b: An open large-scale dataset for training next generation image-text models. NeurIPS  (2022)

\bibitem{schuhmann2021laion}
Schuhmann, C., Vencu, R., Beaumont, R., Kaczmarczyk, R., Mullis, C., Katta, A., Coombes, T., Jitsev, J., Komatsuzaki, A.: Laion-400m: Open dataset of clip-filtered 400 million image-text pairs. arXiv preprint arXiv:2111.02114  (2021)

\bibitem{seo2023let}
Seo, J., Jang, W., Kwak, M.S., Ko, J., Kim, H., Kim, J., Kim, J.H., Lee, J., Kim, S.: Let 2d diffusion model know 3d-consistency for robust text-to-3d generation (2023)

\bibitem{dmtet}
Shen, T., Gao, J., Yin, K., Liu, M.Y., Fidler, S.: Deep marching tetrahedra: a hybrid representation for high-resolution 3d shape synthesis. NeurIPS  (2021)

\bibitem{shi2023mvdream}
Shi, Y., Wang, P., Ye, J., Long, M., Li, K., Yang, X.: Mvdream: Multi-view diffusion for 3d generation (2023)

\bibitem{smith2017improved}
Smith, E.J., Meger, D.: Improved adversarial systems for 3d object generation and reconstruction. In: CoRL (2017)

\bibitem{song2019generative}
Song, Y., Ermon, S.: Generative modeling by estimating gradients of the data distribution. NeurIPS  (2019)

\bibitem{wang2022score}
Wang, H., Du, X., Li, J., Yeh, R.A., Shakhnarovich, G.: Score jacobian chaining: Lifting pretrained 2d diffusion models for 3d generation. CVPR  (2023)

\bibitem{wang2022rodin}
Wang, T., Zhang, B., Zhang, T., Gu, S., Bao, J., Baltrusaitis, T., Shen, J., Chen, D., Wen, F., Chen, Q., Guo, B.: Rodin: A generative model for sculpting 3d digital avatars using diffusion. CVPR  (2023)

\bibitem{wang2022pretraining}
Wang, T., Zhang, T., Zhang, B., Ouyang, H., Chen, D., Chen, Q., Wen, F.: Pretraining is all you need for image-to-image translation (2022)

\bibitem{wang2023prolificdreamer}
Wang, Z., Lu, C., Wang, Y., Bao, F., Li, C., Su, H., Zhu, J.: Prolificdreamer: High-fidelity and diverse text-to-3d generation with variational score distillation (2023)

\bibitem{wu2016learning}
Wu, J., Zhang, C., Xue, T., Freeman, B., Tenenbaum, J.: Learning a probabilistic latent space of object shapes via 3d generative-adversarial modeling. NeurIPS  (2016)

\bibitem{wu2023scoda}
Wu, Y., Yan, Z., Chen, C., Wei, L., Li, X., Li, G., Li, Y., Cui, S., Han, X.: Scoda: Domain adaptive shape completion for real scans. In: CVPR (2023)

\bibitem{xu2023large}
Xu, D., Chen, W., Peng, W., Zhang, C., Xu, T., Zhao, X., Wu, X., Zheng, Y., Chen, E.: Large language models for generative information extraction: A survey. arXiv preprint arXiv:2312.17617  (2023)

\bibitem{xu2023bridging}
Xu, Z., Chen, Z., Zhang, Y., Song, Y., Wan, X., Li, G.: Bridging vision and language encoders: Parameter-efficient tuning for referring image segmentation. ICCV  (2023)

\bibitem{yang2019pointflow}
Yang, G., Huang, X., Hao, Z., Liu, M.Y., Belongie, S., Hariharan, B.: Pointflow: 3d point cloud generation with continuous normalizing flows. In: ICCV (2019)

\bibitem{yang2023llmgrounder}
Yang, J., Chen, X., Qian, S., Madaan, N., Iyengar, M., Fouhey, D.F., Chai, J.: Llm-grounder: Open-vocabulary 3d visual grounding with large language model as an agent (2023)

\bibitem{yu2023mvimgnet}
Yu, X., Xu, M., Zhang, Y., Liu, H., Ye, C., Wu, Y., Yan, Z., Zhu, C., Xiong, Z., Liang, T., et~al.: Mvimgnet: A large-scale dataset of multi-view images. In: CVPR (2023)

\bibitem{zeng2022lion}
Zeng, X., Vahdat, A., Williams, F., Gojcic, Z., Litany, O., Fidler, S., Kreis, K.: Lion: Latent point diffusion models for 3d shape generation. NeurIPS  (2022)

\bibitem{zhang2023text2nerf}
Zhang, J., Li, X., Wan, Z., Wang, C., Liao, J.: Text2nerf: Text-driven 3d scene generation with neural radiance fields (2023)

\bibitem{zhang2023text2layer}
Zhang, X., Zhao, W., Lu, X., Chien, J.: Text2layer: Layered image generation using latent diffusion model. arXiv preprint arXiv:2307.09781  (2023)

\bibitem{zhang2020image}
Zhang, Y., Chen, W., Ling, H., Gao, J., Zhang, Y., Torralba, A., Fidler, S.: Image gans meet differentiable rendering for inverse graphics and interpretable 3d neural rendering. arXiv preprint arXiv:2010.09125  (2020)

\bibitem{zhou20213d}
Zhou, L., Du, Y., Wu, J.: 3d shape generation and completion through point-voxel diffusion. In: ICCV (2021)

\bibitem{zhou2024gala3d}
Zhou, X., Ran, X., Xiong, Y., He, J., Lin, Z., Wang, Y., Sun, D., Yang, M.H.: Gala3d: Towards text-to-3d complex scene generation via layout-guided generative gaussian splatting. arXiv preprint arXiv:2402.07207  (2024)

\end{thebibliography}

\newpage

\section{More Results}
\subsection{Text Splitting.}
\begin{figure}[!htbp]
    \centering
    \includegraphics[width=0.99\linewidth]{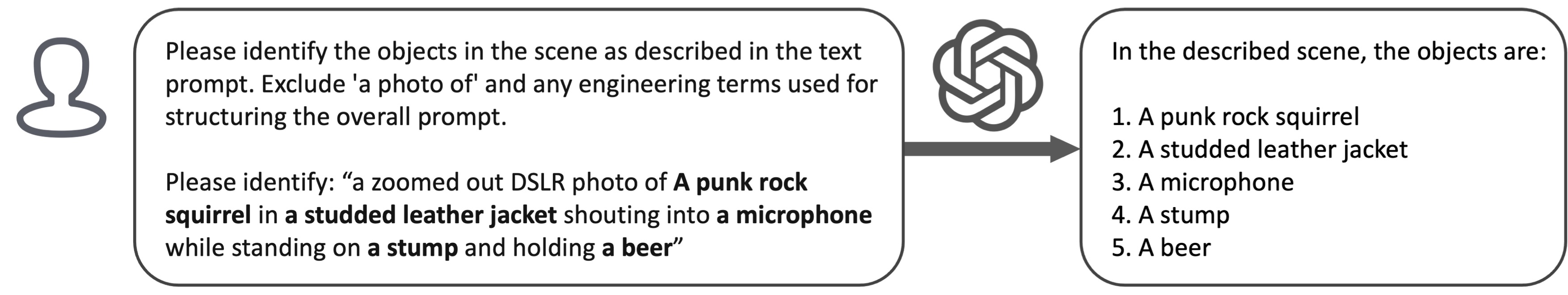}\\
    \caption{
    \textbf{Text splitting.}
    }
    \label{fig:text_splitting}
    \vspace{-4mm}
\end{figure}

The Category Score Distillation Sampling~(CSDS) requires splitting the input text prompts into individual objects. We employ GPT-4 for this purpose, a method commonly used and effective for information extraction~\cite{xu2023large}. We empirically found that GPT-4 has the ability to split very complex text prompts, as shown in Figure~\ref{fig:text_splitting}. 

\subsection{Applications on texture editing}
\begin{figure}[!htbp]
    \centering
    \includegraphics[width=1\linewidth]{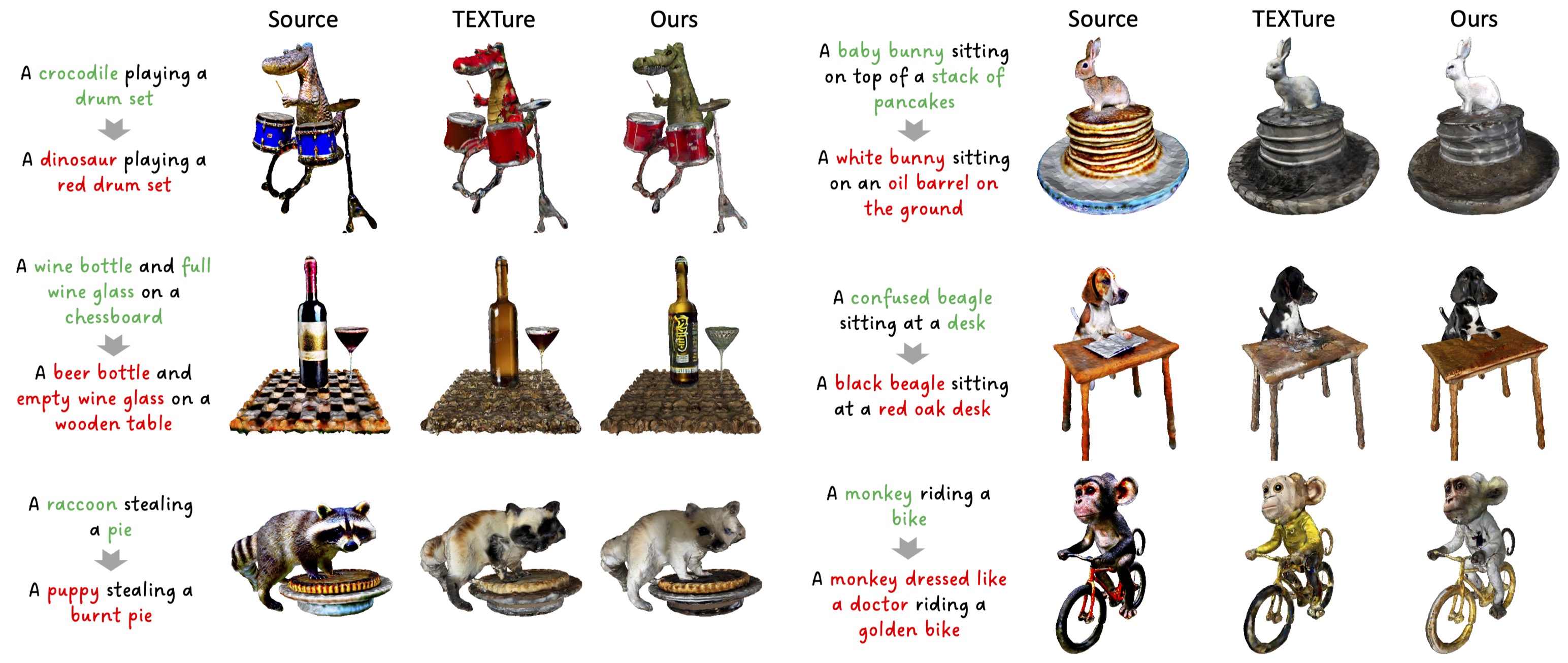}\\
    \caption{
    \textbf{Text-guided texture editing.}
    }
    \vspace{-4mm}
    \label{fig:texture_supp}
\end{figure}

We provide more results on text-guided texture editing, as shown in Figure~\ref{fig:texture_supp}. It can be observed that our method offers greater controllability compared to TEXTure~\cite{richardson2023texture}.

\subsection{Limitations}
\textit{DreamDissector} is likely to fail when objects are in very close contact, such as the body and clothing. We present two examples of this failure in the figure below. The primary reason is the challenge of obtaining clean NeCFs for such complex interactions. 

\begin{figure}[!htbp]
    \centering
    \includegraphics[width=0.75\linewidth]{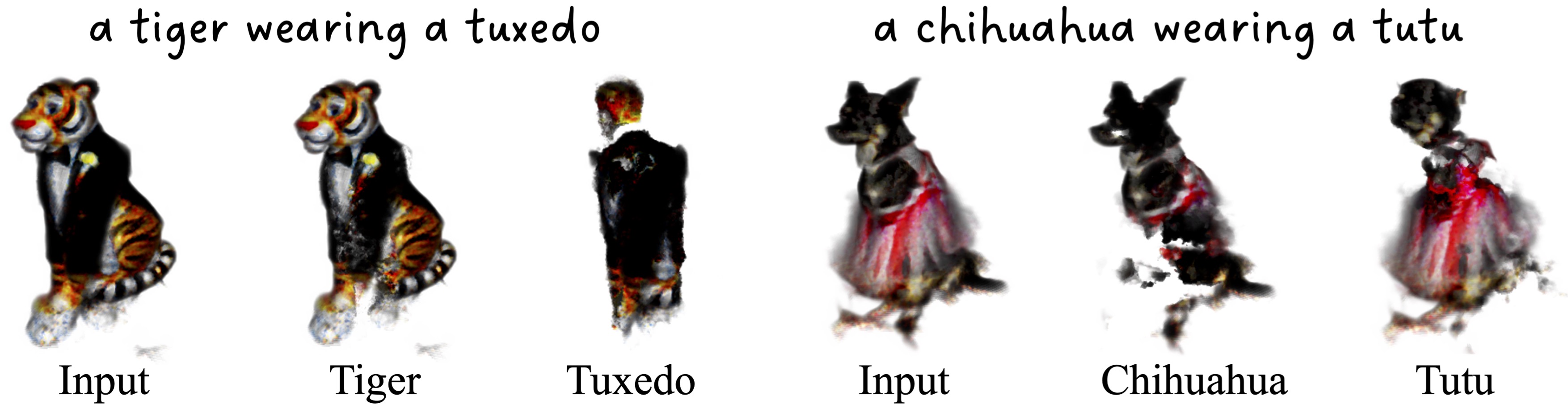}\\
    \vspace{-2mm}
    \caption{
    \textbf{Failure cases.}
    }
    \vspace{-4mm}
    \label{fig:failure_cases}
\end{figure}    
\begin{figure*}[!htbp]
    \centering
    \includegraphics[width=0.99\linewidth]{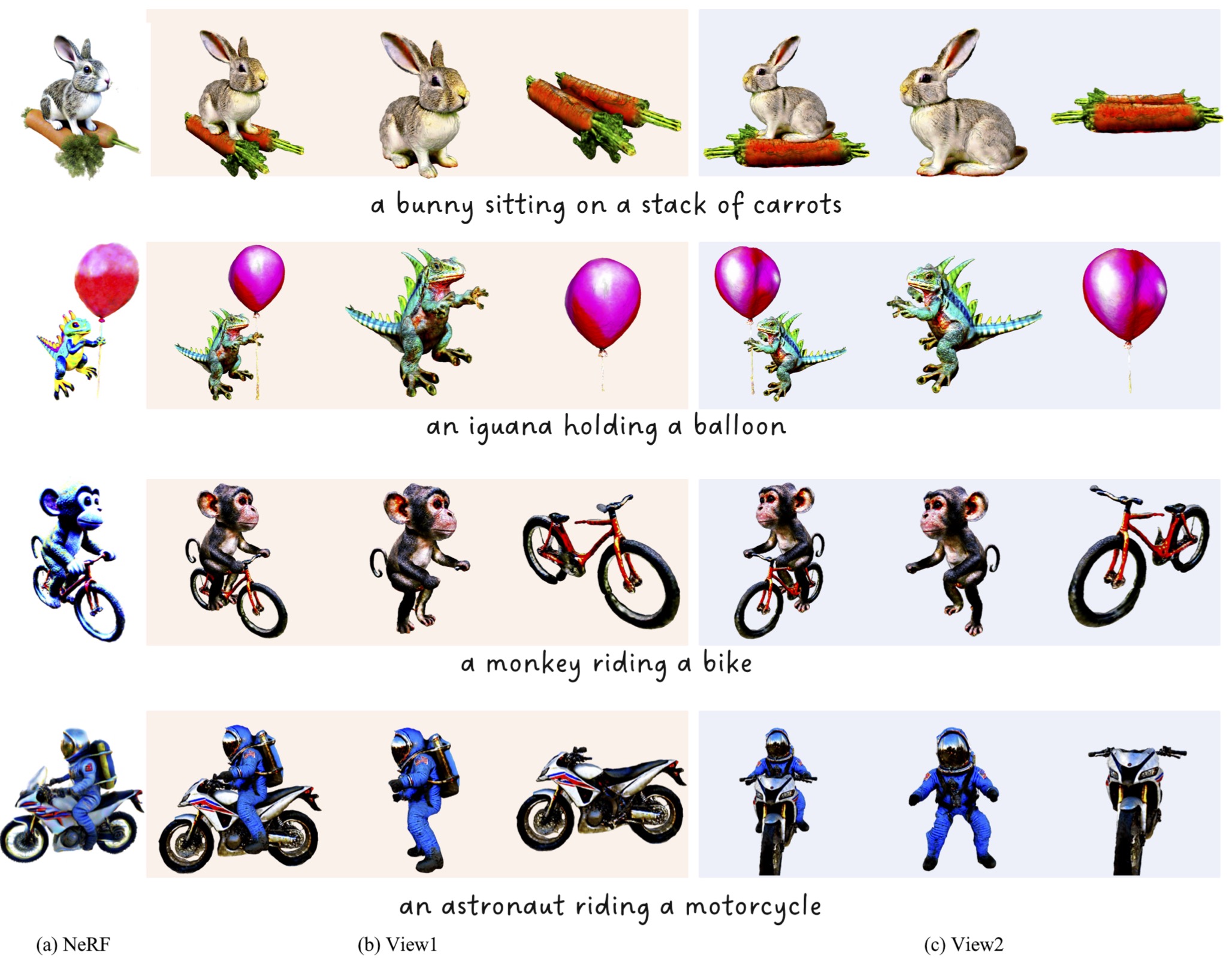}\\
    \caption{\textbf{Qualitative results based on MVDream~\cite{shi2023mvdream}.}
    }
    \label{fig:supp_res_4}
\end{figure*}

\subsection{Results on MVDream}
We adopt Dreamfusion~\cite{poole2022dreamfusion} as the backbone method for generating the initial text-to-3D NeRF for our main results. To verify the versatility of DreamDissector against different backbone methods, we employ MVDream~\cite{shi2023mvdream}, a recently proposed text-to-3D method, as the backbone. Results are shown in Figure~\ref{fig:supp_res_4}. It can be observed that DreamDissector successfully dissects MVDream and produces independent textured meshes with improved geometries and textures.

\subsection{Results on disentangled text-to-3D generation}
\begin{figure*}[!htbp]
    \centering
    \includegraphics[width=0.95\linewidth]{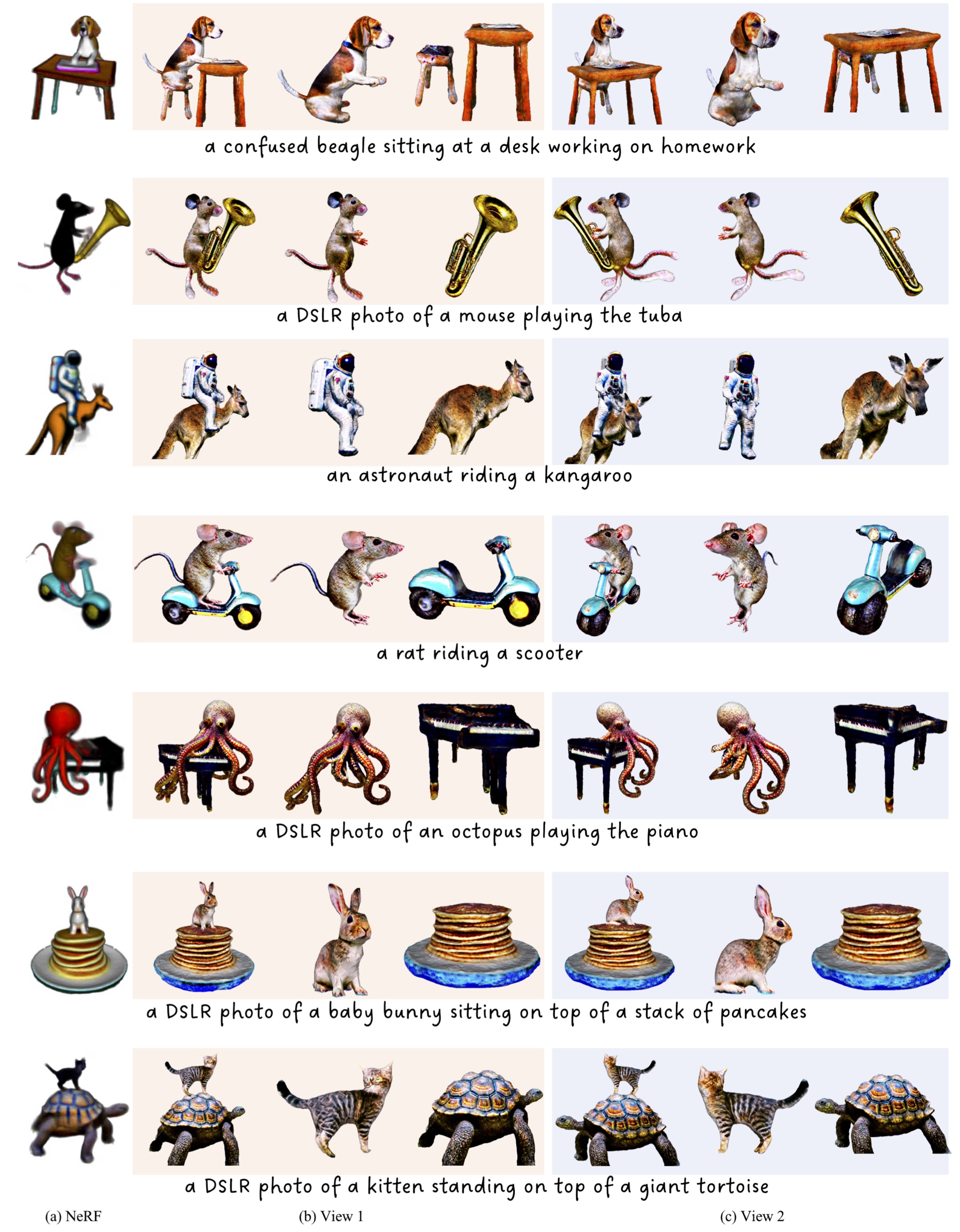}\\
    \caption{\textbf{Qualitative results based on Dreamfusion~\cite{poole2022dreamfusion}.}
    }
    \label{fig:supp_res_1}
\end{figure*}

\begin{figure*}[!htbp]
    \centering
    \includegraphics[width=0.95\linewidth]{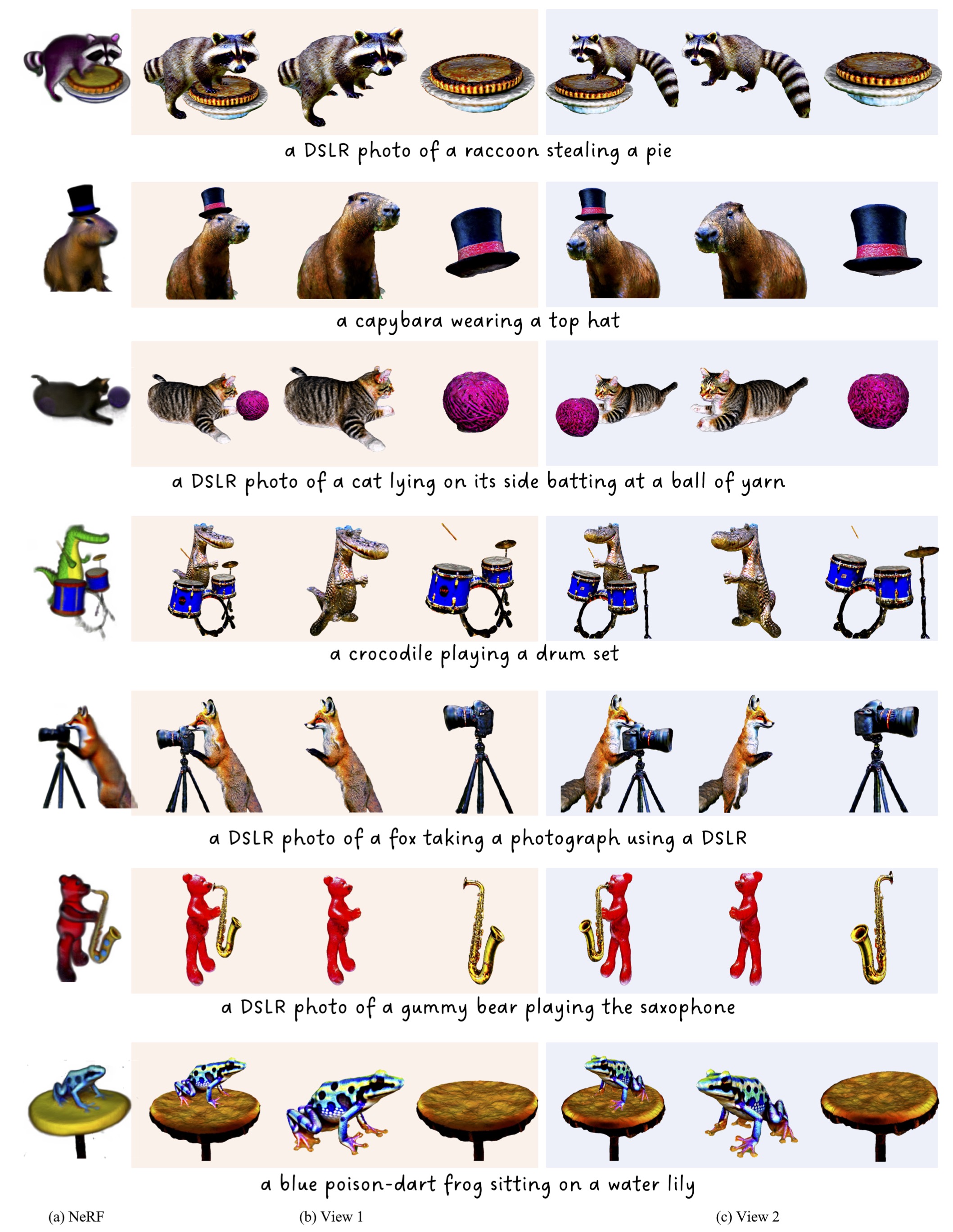}\\
    \caption{\textbf{Qualitative results based on Dreamfusion~\cite{poole2022dreamfusion}.}
    }
    \label{fig:supp_res_2}
\end{figure*}

\begin{figure*}[!htbp]
    \centering
    \includegraphics[width=0.95\linewidth]{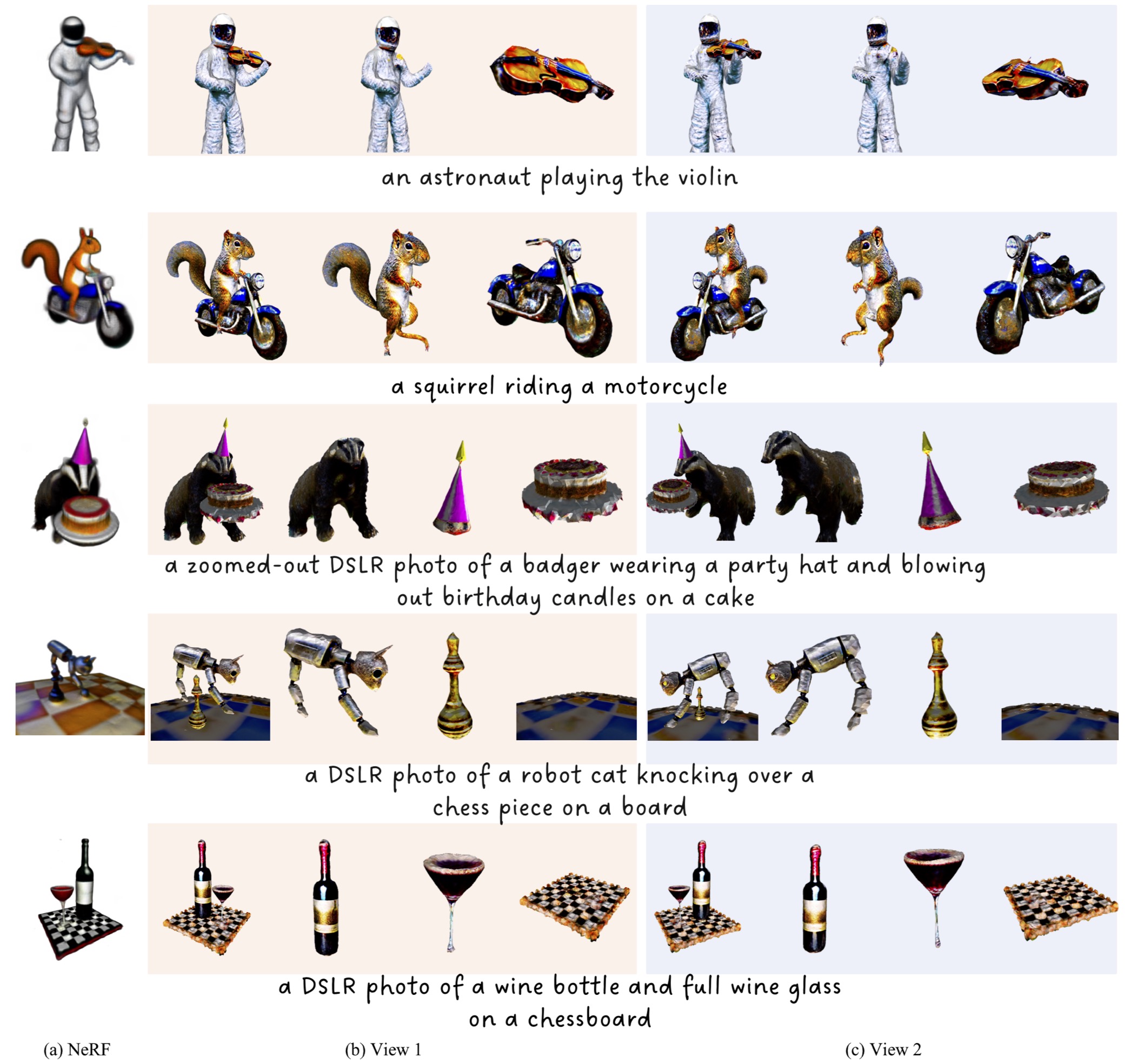}\\
    \caption{\textbf{Qualitative results based on Dreamfusion~\cite{poole2022dreamfusion}.}
    }
    \label{fig:supp_res_3}
\end{figure*}

We present additional results on disentangled text-to-3D generation, including those featured in the main paper. These results and text prompts are depicted in Figure~\ref{fig:supp_res_1},~\ref{fig:supp_res_2} and~\ref{fig:supp_res_3}.

\subsection{Comparisons with the baselines}
\begin{figure*}[!htbp]
    \centering
    \includegraphics[width=0.95\linewidth]{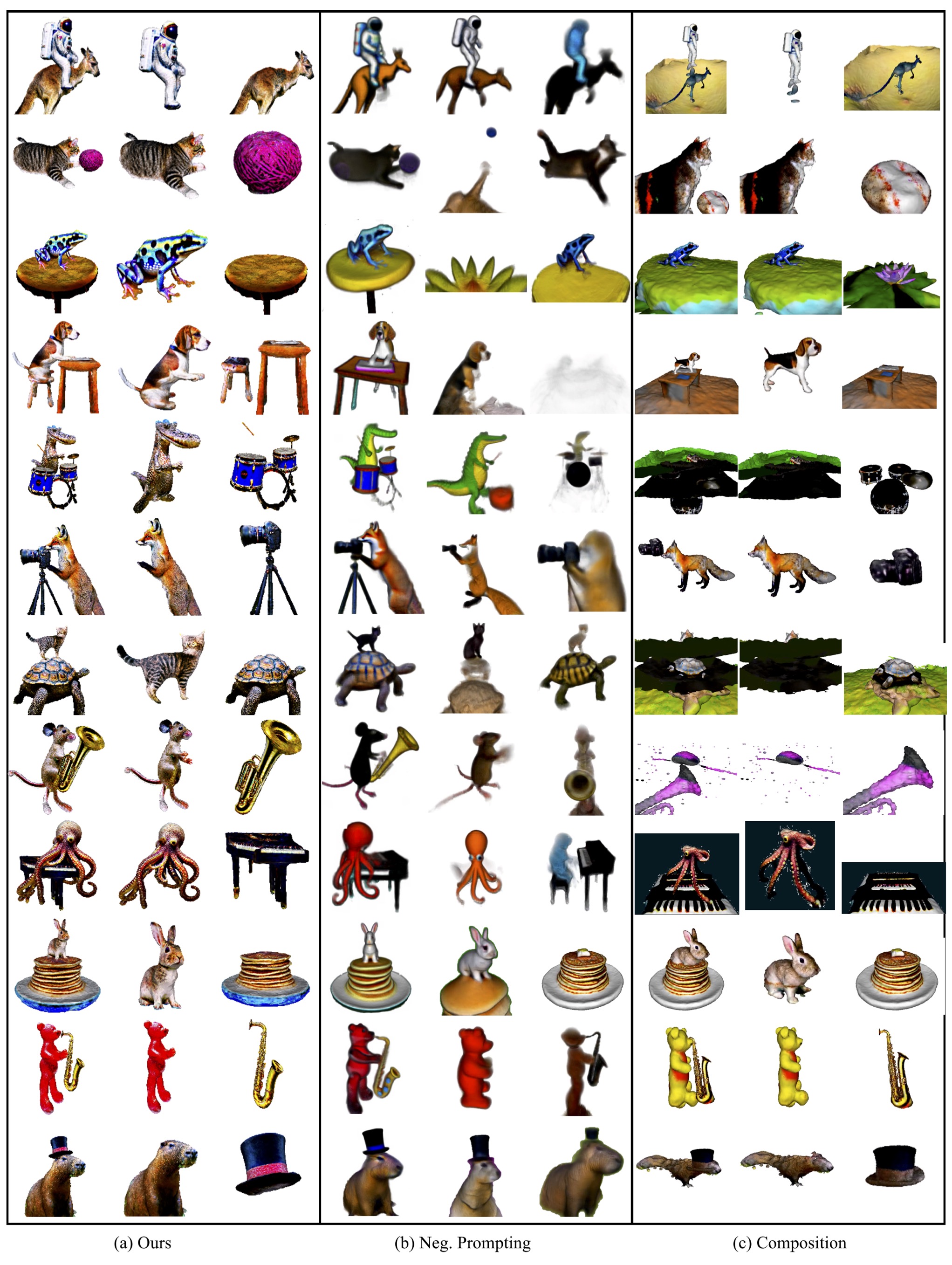}\\
    \caption{\textbf{Comparison with baseline methods.}
    }
    \label{fig:supp_comp}
\end{figure*}

\begin{figure*}[t]
    \centering
    \includegraphics[width=1\linewidth]{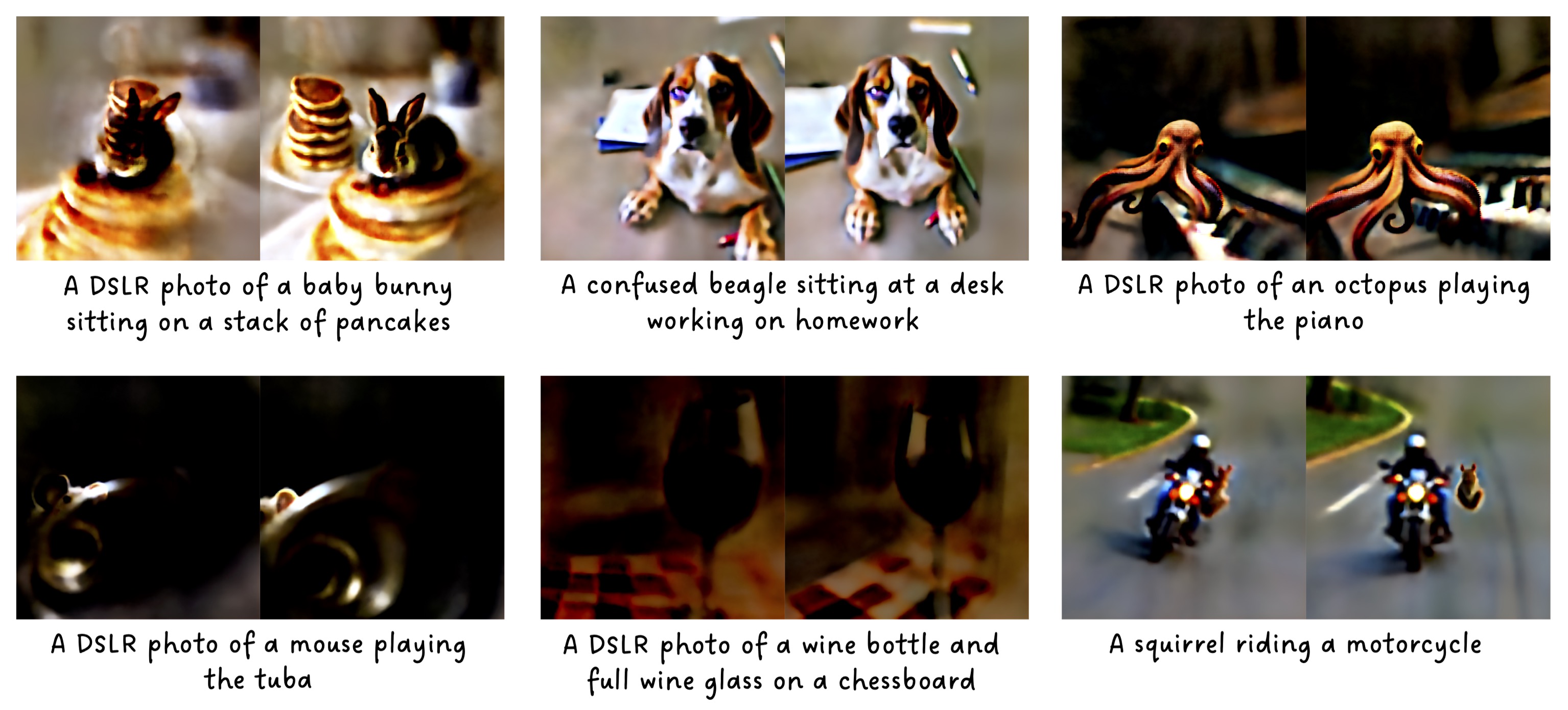}\\
    \caption{\textbf{Results on Set-the-Scene.} We show the results on set-the-scene. It can be observed that set-the-scene struggles to model the object-interected scenes.
    }
    \label{fig:setthescene}
\end{figure*}

Additional comparisons are shown in Figure~\ref{fig:supp_comp}. It should be noted that negative prompting baseline, being intended to generate independent objects, does not associate with composed objects. Therefore, we regard the entire NeRF as the composed object.
We also show the results of a text-guided scene generation method, Set-the-Scene~\cite{set-the-scene}, shown in Figure~\ref{fig:setthescene}. These results illustrate the superior performance of our method.


\end{document}